\documentclass[a4paper,fleqn]{cas-dc}

\usepackage[utf8]{inputenc}
\usepackage{amsmath}
\usepackage{amsthm}
\usepackage{amssymb}
\usepackage{bm}
\usepackage{graphicx}
\usepackage{longtable}
\usepackage{booktabs}
\usepackage{circuitikz}
\usepackage{svg}
\usepackage{pdfpages}
\usepackage{float}
\usepackage[inline,shortlabels]{enumitem}
\usepackage{stfloats}
\usepackage{supertabular}
\usepackage{booktabs}
\usepackage{caption}
\usepackage[numbers, sort&compress]{natbib}
\usepackage{flushend}
\usepackage{tikz}
\usepackage{eurosym} 
\DeclareUnicodeCharacter{20AC}{\euro}
\usepackage{mathtools}
\usepackage{algorithm}
\usepackage{algpseudocode}
\usepackage{natbib}

\newlist{E}{enumerate}{1}
\setlist[E]{label=\textbf{E\arabic*:}}

\usepackage[nameinlink]{cleveref}
\crefname{figure}{Fig.}{Figs.}
\Crefname{figure}{Figure}{Figures}
\crefname{equation}{Eq.}{Eqs.}
\Crefname{equation}{Equation}{Equations}
\crefname{section}{Section}{Sections}
\crefname{table}{Table}{Tables}
\crefname{appendix}{Appendix}{Appendices}
\crefname{algorithm}{Alg.}{Algs.}
\Crefname{algorithm}{Algorithm}{Algorithms}

\newcommand{\eg}{e.g.,\ }

\newcommand{\ie}{i.e.,\ }

\newcommand\norm[1]{\left\lVert#1\right\rVert}

\newcommand{\timestep}{t}

\newcommand{\action}{a}

\newcommand{\price}{\lambda}

\newcommand{\scenarioTree}{\mathcal{T}}
\newcommand{\scenarioTreeNode}{n}
\newcommand{\scenario}{\xi}
\newcommand{\scenarioIndex}{m}
\newcommand{\scenarioProbability}{p}
\newcommand{\scenarioTreeLeaves}{\mathcal{L}}
\newcommand{\scenarioTreeLeafNode}{l}
\newcommand{\numScenarioOriginal}{N_{\text{o}}}
\newcommand{\numScenarioReduced}{N_{\text{r}}}
\newcommand{\scenarioSet}{{\vec{\xi}}}
\newcommand{\scenarioSubset}{\mu}
\newcommand{\splitterFunction}{\$}
\newcommand{\scenarioLength}{H}
\newcommand{\scenarioDimension}{D}
\newcommand{\branchingFactor}{\beta}

\newcommand{\stateSpace}{\mathcal{S}}
\newcommand{\MDPstate}{s}
\newcommand{\actionSpace}{\mathcal{A}}
\newcommand{\transitionFunction}{f}
\newcommand{\rewardFunction}{\rho}
\newcommand{\reward}{r}

\newcommand{\settlementLength}{T}
\newcommand{\priceFormula}{\hat{\price}}
\newcommand{\SI}{x}
\newcommand{\averageSI}{{\hat{\SI}}}
\newcommand{\SIProgressionFormula}{g}
\newcommand{\responseFormula}{h}
\newcommand{\seasonalFactor}{a}
\newcommand{\seasonalFunction}{y}
\newcommand{\balancingFactor}{b}
\newcommand{\stochasticityFactor}{c}
\newcommand{\SIFluctuationsDistribution}{\mathbb{W}}
\newcommand{\SIFluctuations}{w}
\newcommand{\responseSlope}{a_r}
\newcommand{\responseIntercept}{b_r}
\newcommand{\responseLowerBound}{L_r}
\newcommand{\responseUpperBound}{U_r}
\newcommand{\priceSlope}{a_\price}
\newcommand{\priceIntercept}{b_\price}
\newcommand{\priceConstantPositive}{c^+_\price}
\newcommand{\priceConstantNegative}{c^-_\price}
\newcommand{\costExponent}{q}
\newcommand{\timeFactor}{\varphi}
\newcommand{\responseShape}{\bar{h}}
\newcommand{\priceSatLow}{\lambda^{\textsf{lo}}}
\newcommand{\priceSatHigh}{\lambda^{\textsf{hi}}}

\newcommand{\depth}{k}
\newcommand{\leafNode}{\ell}
\newcommand{\budget}{B}

\newcommand{\textualScenarioNode}{scenario }

\begin{document}

\title[mode = title]{S3TS: Stochastic Scenario-Structured Tree Search for Advanced Planning Under Uncertainty}
\shorttitle{S3TS: Stochastic Scenario-Structured Tree Search for Advanced Planning Under Uncertainty}

\author[1]{Fabio Pavirani}[orcid=0009-0005-7904-099X]
\ead{fabio.pavirani@ugent.be}
\credit{Conceptualization, Methodology, Software, Formal analysis, Investigation, Visualization, Writing - Original Draft}

\author[2]{Bert Claessens}[orcid=0009-0006-6116-1483]
\credit{Conceptualization, Methodology, Supervision, Writing - Review \& Editing}

\author[3]{Pierre Pinson}[orcid=0000-0002-1480-0282]
\credit{Conceptualization, Supervision, Writing - Review \& Editing}

\author[1]{Chris Develder}[orcid=0000-0003-2707-4176]
\credit{Supervision, Project administration, Funding acquisition, Writing - Review \& Editing}

\affiliation[1]{organization={IDLab Ghent university -- imec},
                addressline={Technologiepark Zwijnaarde 126}, 
                postcode={9052}, 
                postcodesep={}, 
                city={Gent},
                country={Belgium}}
                
\affiliation[2]{organization={Beebop.ai},
                country={Belgium}}

\affiliation[3]{organization={Imperial College London},
                city={London},
                country={UK},}


\renewcommand{\shortauthors}{Pavirani et al.}

\begin{abstract}
    Effective scheduling in the energy sector is essential to ensure the reliable operation of electrical grids and their connected assets by, for instance, optimizing the dispatch of generation units and storage systems. An effective planning strategy must \begin{enumerate*}[(a)]
      \item accommodate advanced and potentially non-linear system models\,---\,exploiting the increasing data availability of modern grids, and 
      \item explicitly handle uncertainties arising, for instance, from the integration of renewable energy sources. 
    \end{enumerate*}
    While existing approaches can address either non-linearity (e.g., Monte Carlo Tree Search) or uncertainty (e.g., stochastic mathematical optimization), there is a lack of planning techniques capable of addressing both challenges simultaneously.
    To bridge this gap, we propose a Stochastic Scenario-Structured Tree Search (S3TS) algorithm that explicitly represents uncertainty through scenario trees while enabling the integration of advanced non-linear models. We evaluate S3TS on a simulated demand response signal publication problem, largely mimicking the imbalance settlement mechanism in Belgium. The results demonstrate near-optimal performance in linear, analytically tractable settings, with costs within 14\% of the mathematically optimal solution conditioned to the scenario trees. In highly non-linear scenarios, S3TS significantly outperforms baseline methods, achieving cost reductions of up to 51\% and 5.4\% compared to a myopic algorithm and deterministic MCTS, respectively.
\end{abstract}

\maketitle

\section{Introduction}
The growing complexity of electrical power systems, driven by the widespread electrification of energy sources, has made effective planning strategies essential for ensuring reliable grid operation. 
Accurate forecasting and strategic decision-making enable both grid operators and grid users to effectively optimize their objectives~\cite{zahraoui2025assessment}. For instance, grid operators such as Transmission System Operators (TSOs) can forecast the grid load and imbalance, and plan balancing activations accordingly to favor grid stability. Similarly, grid users (such as Balance Responsible Parties (BRPs), residential, and commercial consumers) benefit from accurate forecasts of future energy prices, which allow them to adjust their consumption in line with expected grid conditions. This mechanism, also known as Demand Response (DR)~\cite{sezgen2007option}, is significantly contributing to grid stability~\cite{silva2022demand}, and relies on effective planning by both operators and users.

A widely adopted approach for such planning tasks is mathematical optimization, particularly in the form of Model Predictive Control (MPC)~\cite{bakirtzis2012generation,he2025efficient}. 
MPC formulates sequential control tasks as optimization problems, aiming to minimize expected costs associated with controllable actions across the prediction horizon. 
This requires a dynamic model of the system to simulate state evolution and cost variations under different actions. 
Once an optimal solution has been found (usually through mathematical solvers), the resulting optimal actions are applied in a receding-horizon fashion, \ie only the first action is executed before the optimization is repeated at the next time step.

Yet, as renewable energy sources become more prevalent, the resulting stochasticity in system dynamics poses a fundamental challenge to effective planning.
In fact, if these uncertainties are not properly addressed, planning strategies can become fragile and unstable, potentially leading to substantial economic and operational losses~\cite{haugen2023representation,roald2023power}. 
To mitigate this, multi-stage stochastic optimization methods have been used~\cite{birge1997introduction}. 
These methods represent uncertainty through scenario trees constructed from sampled realizations of the underlying random variables. The scenario tree provides a structured representation of how uncertainty may evolve over time, with each branch corresponding to a possible future outcome. Decisions are then optimized across these scenarios using mathematical programming techniques.
Stochastic MPC is a prominent example of this approach. Example of its application can be found for both grid operators~\cite{patrinos2011stochastic,lei2021multi} and grid users~\cite{smets2023strategic,madahi2025model}.

Despite their popularity, mathematical optimization methods face two major challenges. 
First, accurate modeling of system dynamics is critical for effective planning. 
As power systems grow more complex, increasingly advanced models with highly non-linear dynamics are typically required. 
Integrating such models into mathematical optimization frameworks is difficult due to their non-linear dynamics~\cite{roald2023power}. 
Second, evolving market designs introduce empirical rules and constraints that are hard to model analytically within conventional optimization techniques~\cite{anderson2025ten,kotzur2021modeler}. 
While simplified approximations can be used, they often fail to reflect the true complexity of modern power systems, resulting in suboptimal strategies.

These challenges thus call for planning techniques that can handle both non-linear models and complex market structures without sacrificing control robustness. 
Monte Carlo Tree Search (MCTS) has emerged as a promising alternative in the energy field, applied to problems such as residential heating~\cite{pavirani2024demand,kiljander2021intelligent}, imbalance price publication~\cite{pavirani2025predicting}, complex system operation and maintenance~\cite{hao2024monte}, and microgrid energy scheduling~\cite{shuai2020online}.
MCTS explores the decision space by simulating multiple action sequences (trajectories) and selecting the most promising ones. 
Unlike mathematical optimization, MCTS imposes no structural constraints on the models beyond their ability to be simulated. 

Standard MCTS, however, assumes deterministic transitions, which limits its applicability when uncertainty plays a dominant role. Several extensions have been proposed in the tree search literature to relax this assumption. The classical solution, originating in the game-tree community, is to interleave \emph{decision nodes} with \emph{chance nodes} in the search tree: at each chance node, an outcome is sampled from a (typically known) transition distribution, and the value backed up is the expectation over outcomes\,---\,\eg in $*$-minimax~\cite{ballard1983minimax} for stochastic games. \citet{kearns2002sparse} formalized this idea for general MDPs through their sparse-sampling algorithm, which builds a finite-depth tree by drawing a fixed number of next-state samples per action, providing near-optimal value estimates with sample complexity independent of the state-space size. POMCP~\cite{silver2010monte} extended chance-node MCTS to partially observable MDPs by sampling particles from the belief state at every simulation, while continuous-state stochastic dynamics are typically handled through double progressive widening~\cite{couetoux2011continuous}, which gradually expands new outcome children as the visit count of a chance node grows. Although these methods cover a broad spectrum of stochastic settings, they share a common design choice: uncertainty is sampled online during simulation, drawing fresh outcomes from a transition oracle each time a chance node is visited. This couples the representation of uncertainty to the search procedure and presupposes cheap access to such an oracle, neither of which aligns naturally with the multi-stage stochastic programming tradition\,---\,where uncertainty is instead encoded upfront in a structured, finite scenario tree obtained through probabilistic forecasting and scenario reduction.
 
A more recent line of work tackles the deterministic-dynamics limitation by \emph{learning} a stochastic transition model end-to-end. Building on MuZero~\cite{schrittwieser2020mastering}, \citet{antonoglou2021planning} proposed Stochastic MuZero, where a variational autoencoder produces discrete latent outcomes that are then expanded as chance children during planning. While powerful in domains with abundant interaction data, this approach inherits two limitations relevant to our application: the latent transition model is opaque, and it is decoupled from the substantial body of work on scenario generation and reduction developed in stochastic programming~\cite{heitsch2003scenario,growe2003scenario,heitsch2009scenario,pinson2009probabilistic}, both of which are obstacles to adoption by real-world energy operators who require auditable forecasts. Our proposal sits at the intersection of these two traditions. Rather than sampling outcomes online from a transition oracle (as in chance-node MCTS, sparse sampling, or POMCP), or learning them implicitly via a latent model (as in Stochastic MuZero), we consider a \emph{pre-built scenario tree} as input and use its topology to structure the search. Concretely, this differs from a chance-node MCTS along two axes:
\begin{enumerate*}[(i)]
    \item the branching topology at every stochastic split is fixed externally by the scenario tree, hence no resampling occurs at visit time and the value estimates do not carry the variance of online Monte Carlo draws; and
    \item the probabilities attached to each branch come from a scenario reduction step (\eg~\cite{heitsch2003scenario}) rather than from a sampling oracle, which makes the framework compatible with any probabilistic forecasting pipeline.
\end{enumerate*}
A practical corollary is that the same scenario tree used for stochastic MPC is also used for our proposed tree search algorithm, enabling controlled comparisons that isolate the contribution of the optimization procedure from that of the uncertainty representation\,---\,a property we exploit in the experimental evaluation.
The net effect is to bring the trustworthiness and modeling discipline of multi-stage stochastic programming into the flexibility of MCTS-style search.

Building on this positioning, we introduce \emph{Stochastic Scenario-Structured Tree Search} (S3TS) and use it to bring the modeling discipline of multi-stage stochastic programming into MCTS-style search. The flexibility inherited from MCTS makes S3TS particularly attractive in settings where system dynamics are better captured through non-linear data-driven models\,---\,for instance, residential heating systems with neural-network-based thermal models~\cite{pavirani2024demand,kiljander2021intelligent}\,---\,or where the cost function contains components that hinder conventional optimization, such as discontinuous price ladders, $\min$/$\max$ operators, or exponential terms combined with mixed-integer constraints. Anchoring the search to a pre-built scenario tree, in turn, aligns the resulting planner with established stochastic-programming pipelines: the same scenario generation, scenario reduction, and tree assembly steps already used to feed a stochastic MPC can directly drive S3TS, without modification.
The key contributions of our work are:
\begin{enumerate}
    \item We introduce S3TS (\cref{sec:S3TS_methodology}), an extension of MCTS that incorporates uncertainty through an externally constructed scenario tree rather than through online sampling from a transition oracle or a learned latent model. This makes the search topology, branch probabilities, and depth all explicit and inspectable, and exposes the same scenario-tree interface used by classical multi-stage stochastic optimization (as discussed in \cref{sec:methodology}).
    \item We formalize a Demand Response (DR) signal publication problem\,---\,inspired by the imbalance settlement mechanism currently in place in Belgium~\cite{elia2024imbalance}\,---\,in both a linear and a non-linear formulation (\cref{sec:problem_description}), and solve the linearized version through Mixed-Integer Linear Programming to provide a reference upper bound.
    \item We empirically evaluate S3TS on this problem (\cref{sec:experimental_setup}). Our results (\cref{sec:results}) show that in the linear formulation, S3TS achieves a total cost within 14\% of the mathematically optimal solution obtained from the same scenario tree, demonstrating that the algorithm closely tracks the optimum even without direct access to the analytical model. In the non-linear formulation\,---\,intractable for conventional solvers\,---\,S3TS reduces cost by 5.4\% relative to deterministic MCTS and by 51\% relative to a rule-based baseline, demonstrating the benefit of jointly handling non-linear dynamics and explicit uncertainty.
\end{enumerate}
\Cref{sec:conclusions} closes with concluding remarks and directions for future work.

\section{Problem Description}
\label{sec:problem_description}
To evaluate our S3TS algorithm, we apply it to a simulated DR market price (signal) publication problem.
First, we evaluate S3TS in a simplified formulation of the problem (\cref{sec:linear_setting_env}). 
Because this problem exhibits linear dynamics, the optimal stochastic solution can be computed analytically. 
This allows us to benchmark S3TS against an upper bound provided by the true optimal solution, offering a clear measure of its efficiency.
Then, we expand the problem by including highly non-linear dynamics (\cref{sec:non-linear_setting_env}), making it analytically intractable for conventional mathematical optimization techniques. We use this second formulation to assess the S3TS performance when applied to more complicated environments.
Since the optimal solution is then unknown, we compare S3TS against two baselines: 
\begin{enumerate*}[(i)]
    \item its deterministic counterpart and
    \item a myopic rule-based strategy, 
\end{enumerate*}
highlighting the advantages of S3TS in complex, uncertainty-rich environments.

\subsection{Price Publication Problem}
\label{sec:price_setting_env}
We consider a problem formulation inspired by the imbalance settlement mechanism as operated in Belgium~\cite{elia2024imbalance,pavirani2025predicting}.\footnote{Note that the mechanism dynamics we considered are not unique to the Belgian system, but it is rather generalizable to other European countries too}
We assume a cooperative setting between two types of parties:
\begin{itemize}
    \item The \textit{system}, whose objective is to maintain its power balance.
    \item A group of \textit{actors}, whose objective is to maximize their economic revenues.
\end{itemize}
The system health is described by its System Imbalance (SI) (\ie the excess/lack of power in the system at a certain instant), and the system's goal would be to keep the SI magnitude as low as possible.
While the actors can influence the state of the system by changing their energy production/consumption, their participation is subject to prices determined by the SI. These prices are computed at the end of each settlement period\,---\,each lasting $\settlementLength \in \mathbb{N}^+$ timesteps (\eg $\settlementLength \doteq 15$ minutes in Belgium)\,---\,using a pricing formula based on the average SI observed during that period.
The system benefits when actors adjust their actions to reduce SI; therefore, the price formula is designed to make helpful actors' influence (\ie actors' actions that reduce the SI magnitude) profitable for them. 
However, since prices are only calculated after the settlement period, actors face uncertainty: their interventions might lead to financial losses if the final price does not match their expectations. In other words, if the SI significantly swings within a settlement period, the actors might struggle to understand which actions would be beneficial for the system; thus, potentially facing financial losses. 
This uncertainty discourages active participation.
To incentivize actors, the system can publish an approximation of the final price \emph{within} the settlement period.
The more accurate this approximation, the lower the risk for actors, making them more willing to participate and influence the SI with the goal of financial gains. 
This creates a mutual benefit: actors face less risk while the system improves its balance thanks to the more active actors' participation. Therefore, publishing accurate prices within the running settlement period is an important problem to enable effective cooperation between the two parties.

Analytically speaking, the problem of publishing accurate prices (signals) can be formulated as follows
{\setlength{\mathindent}{0pt}
\begin{small}
\begin{align}
\label{eq:deterministic_price_publication_problem}
    \min_{\{ \price_{\timestep} \}_{\timestep = 0, 1, \dots, \settlementLength-1}} \quad & \sum_{\tau = 0}^{\settlementLength-1} { {\norm{ \price_{\tau} - \hat{\price}_{\settlementLength} }}_{\costExponent} } \\
    \text{s.t.}\quad
    & \hat{\price}_{\settlementLength} = \priceFormula(\averageSI_{\settlementLength}) , \nonumber\\
    & \SI_{\timestep} = \SIProgressionFormula(\SI_{\timestep-1}) + \responseFormula(\price_{\timestep}) & \forall\, \timestep = 0, 1, \dots, \settlementLength \nonumber\\
    & \SIProgressionFormula(\SI_{\timestep-1}) = \seasonalFactor \,\seasonalFunction(\timestep) + (1-\balancingFactor) \, \SI_{\timestep-1} + \stochasticityFactor \, \SIFluctuations_{\timestep} & \forall\, \timestep = 1, \dots, \settlementLength  \nonumber
\end{align}
\end{small}
}
where:
\begin{itemize}
    \item $\price_{\timestep}$ is the published price (signal) at timestep $\timestep$,
    \item $\hat{\price}_{\settlementLength}$ is the real price that will be charged to the actors, as determined at the end of the settlement timeslot,
    \item $\costExponent \in \mathbb{R}^+$ defines the polynomial cost of the price inaccuracies, 
    \item $\priceFormula(\averageSI_{\settlementLength-1})$ is the function used to calculate the final price based on the averaged measured SI,
    \item $\SI_{\timestep}$ is the SI measured at timestep $\timestep$,
    \item $\averageSI_{\settlementLength}$ is the average SI measured in the timesteps from $0$ to $\settlementLength$, 
    \item $\SIProgressionFormula(\SI_{\timestep-1})$ describes the general trend of the SI, and it comprises 
    \begin{enumerate*}[(i)]
        \item a \emph{seasonal progression} $\seasonalFunction(\timestep) : \mathbb{N} \rightarrow \mathbb{R}$ with relevance $\seasonalFactor \in \mathbb{R}^+$,
        \item a \emph{balancing factor} $\balancingFactor \in [0, 1]$, that describes the general trend of the SI to tend toward 0, and
        \item a \emph{stochastic fluctuation} part $\SIFluctuations_{\timestep}$ with a relevance factor $\stochasticityFactor \in \mathbb{R}^+$ and sampled from a random variable distribution $\SIFluctuationsDistribution_{\timestep}$.
    \end{enumerate*}
    Notice that this function is independent from the published prices\,---\,\ie this describes the SI progression if the actors do not manifest their influence on the system.
    \item $\responseFormula(\price_{\timestep})$ is the influence that the actors manifest over the system based on the published price (signal). This function can potentially be highly non-linear, as it describes the strategy that the actors follow to maximize their revenues (\eg rule-based, empirical strategies, RL-based strategies).
\end{itemize}

Notice that the addition of $\responseFormula(\price_{\timestep})$ potentially makes the problem non-tractable for a conventional optimization algorithm, due to the dependence of the function on the controllable variable ($\price_{\timestep}$). If the function has a highly non-linear structure, the overall problem becomes hardly solvable for a mathematical solver, hindering the final solution. 
In the deterministic formulation defined above, the values of the stochastic variable $\SIFluctuations_{\timestep}$ will be represented by the forecasted expected value of the distribution $\SIFluctuationsDistribution_{\timestep}$. 

In its stochastic formulation, the problem would also have a scenario tree $\scenarioTree$ associated, describing possible future SI fluctuations outcomes sampled from the distribution functions $\SIFluctuationsDistribution_{\timestep}$. 
The nodes of the scenario tree are indicated using the subindex notation and the symbol $\scenarioTreeNode$. 
In this formulation, the decisions of the optimization problem happen on a node level, rather than on a time level. 
Therefore, the resulting optimization problem would be: 
{\setlength{\mathindent}{0pt}
\begin{small}
\begin{align}
\label{eq:stochastic_price_publication_problem}
    \min_{\{ \price_{\scenarioTreeNode} \}_{\scenarioTreeNode \in \scenarioTree \setminus \scenarioTreeLeaves}} \quad & \sum_{\scenarioTreeLeafNode \in \scenarioTreeLeaves}{ \sum_{\scenarioTreeNode' \prec \scenarioTreeLeafNode}{ \mathbb{P}(\scenarioTreeNode') {\norm{ \price_{\scenarioTreeNode'} - \hat{\price}_{\scenarioTreeLeafNode} }}_{\costExponent} } } \\
    \text{s.t.}\quad
    & \hat{\price}_{\scenarioTreeLeafNode} = \priceFormula(\averageSI_{\scenarioTreeLeafNode}) , \nonumber\\
    & \SI_{\scenarioTreeNode} = \SIProgressionFormula(\SI_{\text{par}(\scenarioTreeNode)}) + \responseFormula(\price_{\text{par}(\scenarioTreeNode)}) & \forall\, \scenarioTreeNode \in \scenarioTree \setminus \{0\} \nonumber\\
    & \SIProgressionFormula(\SI_{\text{par}(\scenarioTreeNode)}) = \seasonalFactor \seasonalFunction(\timestep_{\scenarioTreeNode}) + (1-\balancingFactor)\SI_{\text{par}(\scenarioTreeNode)} + \stochasticityFactor \SIFluctuations_{\scenarioTreeNode} & \forall\, \scenarioTreeNode \in \scenarioTree \setminus \{0\}  \nonumber
\end{align}
\end{small}}where $\scenarioTreeLeaves$ is the set of leaves in the scenario tree, $\text{par}(\scenarioTreeNode)$ indicates the parent node of $\scenarioTreeNode$, $\scenarioTreeNode' \prec \scenarioTreeLeafNode$ is used to indicate all the nodes $\scenarioTreeNode'$ preceding (having a parental relation with) $\scenarioTreeLeafNode$, $0$ indicates the root node, and $\timestep_{\scenarioTreeNode}$ is the timestep associated with the node $\scenarioTreeNode$. In this formulation, we assume the leaf nodes correspond temporally to the beginning of the next settlement period (and therefore to the end of the current one); consequently, their state determines the final imbalance price. 

\subsubsection{Linear formulation}
\label{sec:linear_setting_env}
To allow commercial mathematical solvers to solve the problems described in \cref{eq:deterministic_price_publication_problem,eq:stochastic_price_publication_problem}, we can define the functions $\responseFormula: \mathbb{R}\rightarrow\mathbb{R}$ and $\hat{\price}: \mathbb{R}\rightarrow\mathbb{R}$ with a `\emph{low grade of non-linearity}'; in other words, we can ensure that those functions will allow a computationally affordable mathematical solution of the optimization problem by using standard mathematical solvers. 
Doing so allows us to obtain a mathematical bound on the obtainable objective given a certain scenario tree as input. This mathematical bound will give a clear measure of the S3TS efficacy. 
To allow for such a solution, we first considered a linear problem formulation:
\begin{align}
    \responseFormula(\price) &=
        \begin{cases}
            \responseLowerBound, & \text{if } \price < \frac{\responseLowerBound-\responseIntercept}{\responseSlope}, \\
            \responseSlope \price + \responseIntercept, & \text{if } \frac{\responseLowerBound-\responseIntercept}{\responseSlope} \leq \price \leq \frac{\responseUpperBound-\responseIntercept}{\responseSlope}, \\
            \responseUpperBound, & \text{if } \price > \frac{\responseUpperBound-\responseIntercept}{\responseSlope}.
        \end{cases} \\
    \priceFormula(x) &=
        \begin{cases}
            \priceSlope x + \priceIntercept + \priceConstantPositive, & \text{if } x < 0, \\
            \priceSlope x + \priceIntercept + \priceConstantNegative, & \text{if } x > 0, \\
        \end{cases}
\end{align}
where ${\responseSlope < 0, \responseIntercept \in \mathbb{R}},\; {\responseLowerBound \in \mathbb{R}, \responseUpperBound \in \mathbb{R}}$ with $\responseLowerBound < \responseUpperBound$ define a saturated linear response function, and ${\priceSlope < 0},\, {\priceIntercept,\priceConstantPositive,\priceConstantNegative \in \mathbb{R}}$ define a linear price formula with a discontinuous jump when the SI sign changes.
With these definitions, both optimization problems described in \cref{eq:deterministic_price_publication_problem,eq:stochastic_price_publication_problem} are solvable with a standard mathematical solver such as Gurobi~\cite{gurobi}. In this linear formulation, the polynomial cost term used in \cref{eq:deterministic_price_publication_problem,eq:stochastic_price_publication_problem} is fixed to $\costExponent \doteq 1$ to make the optimization solvable with a Mixed-Integer Linear Programming algorithm. 

\subsubsection{Non-Linear formulation}
\label{sec:non-linear_setting_env}
 
The linear formulation in \cref{sec:linear_setting_env} is intentionally simple: it admits a tractable MILP encoding and provides a yardstick for measuring the optimality gap of S3TS. It does not, however, reflect several characteristics of real imbalance settlement mechanisms\,---\, \eg actual price functions are tabulated rather than affine, actors' responsiveness varies during the settlement period, and penalties on inaccurate publications are potentially polynomial rather than linear. To assess the benefits of S3TS over deterministic alternatives in a more realistic regime, we therefore introduce a second formulation that incorporates these features. The added non-linearities make the problem intractable for standard mathematical solvers, while remaining tractable for S3TS, which only requires $\priceFormula$ and $\responseFormula$ to be evaluable in simulation.
 
\begin{figure}
    \centering
    \includegraphics[width=0.5\textwidth]{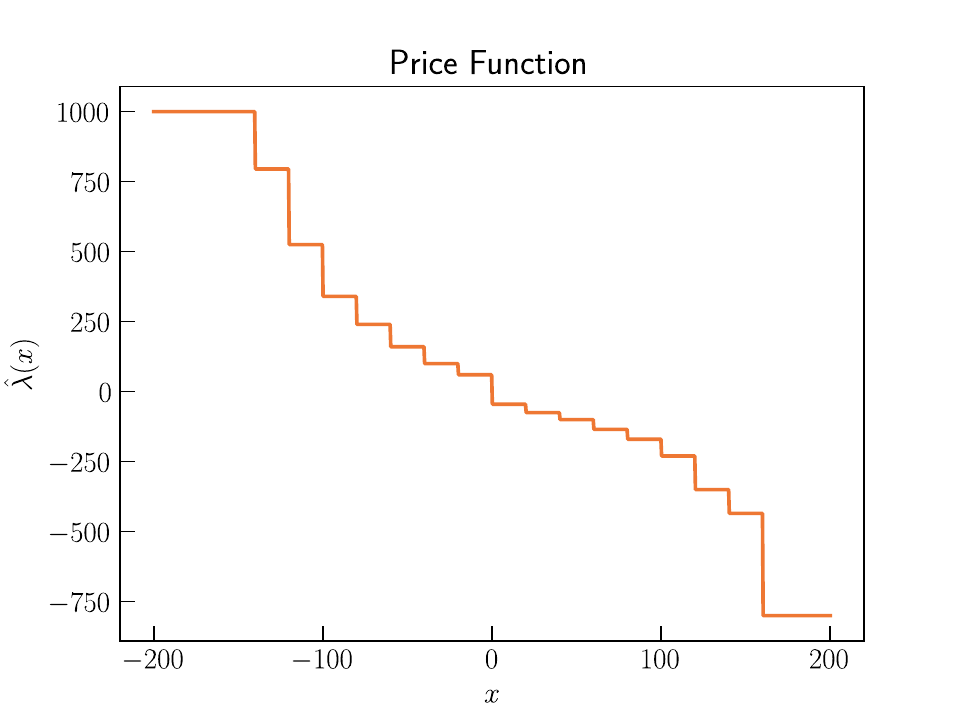}
    \caption{Price function $\priceFormula(\SI)$ used in the non-linear problem formulation.}
    \label{fig:price_function}
\end{figure}

We model $\priceFormula(\SI)$ as a discontinuous step ladder, plotted in \cref{fig:price_function}. The ladder is inspired by the structure currently in place in the Belgian imbalance settlement~\cite{elia2024imbalance}: price magnitude grows super-linearly with $|\SI|$, with steep penalties at large imbalance, and the function exhibits an asymmetric discontinuity at $\SI = 0$ separating the small-positive-$\SI$ regime (mildly negative price) from the small-negative-$\SI$ regime (mildly positive price). The asymmetry around zero reflects the asymmetric costs of upward and downward regulation that are characteristic of real-world balancing markets.

\begin{figure}
    \centering
    \includegraphics[width=0.5\textwidth]{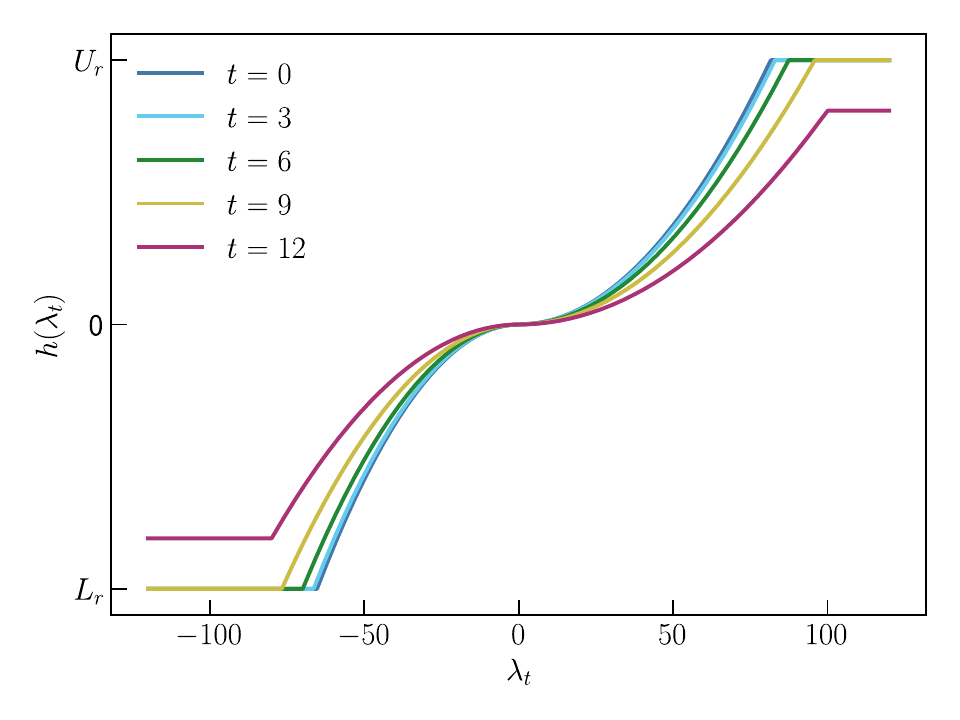}
    \caption{Response function $\responseFormula(\price;\timestep)$ describing the actors' influence over the system in the non-linear problem formulation, plotted at five values of $\timestep$ within a settlement period. Curves at different $\timestep$ correspond to different temporal scalings $\timeFactor(\timestep)$, leading to different effective steepness of the spatial shape.}
    \label{fig:response_function}
\end{figure}

The actors' influence $\responseFormula(\price; \timestep)$ is modeled as the product of a spatial shape and a temporal scaling, clipped to a bounded range:
\begin{equation}
\label{eq:nonlinear_response}
    \responseFormula(\price; \timestep) \;\doteq\; -\,\mathrm{clip}\!\Bigl(\,\timeFactor(\timestep)\;\responseShape(\price),\; \responseLowerBound,\; \responseUpperBound\,\Bigr),
\end{equation}
where the spatial shape $\responseShape : \mathbb{R} \to \mathbb{R}$ is the quadratic-saturated function
\begin{equation}
\label{eq:nonlinear_response_shape}
    \responseShape(\price) =
    \begin{cases}
        \responseUpperBound, & \price < \priceSatLow, \\[2pt]
        \responseUpperBound\,\Bigl(\tfrac{\price}{\priceSatLow}\Bigr)^{\!2}, & \priceSatLow \le \price < 0, \\[2pt]
        \responseLowerBound\,\Bigl(\tfrac{\price}{\priceSatHigh}\Bigr)^{\!2}, & 0 \le \price \le \priceSatHigh, \\[2pt]
        \responseLowerBound, & \price > \priceSatHigh,
    \end{cases}
\end{equation}
and the temporal factor $\timeFactor : \mathbb{N} \to \mathbb{R}^+$ is
\begin{equation}
\label{eq:nonlinear_time_factor}
    \timeFactor(\timestep) \;\doteq\; \Bigl|\cos\!\bigl(2\pi\,(\timestep \bmod \settlementLength)/60\bigr)\Bigr| + \tfrac{1}{2}.
\end{equation}
Here $\priceSatLow < 0 < \priceSatHigh$ are the price thresholds beyond which the actors' response saturates, and $\responseUpperBound,\responseLowerBound \in \mathbb{R}^+$ are the maximum and minimum response magnitude, respectively. The $\mathrm{clip}$ operator in \cref{eq:nonlinear_response} enforces the bound $|\responseFormula| \le \responseUpperBound$ even when $\timeFactor(\timestep) > 1$. The function is depicted at five values of $\timestep$ in \cref{fig:response_function}.
We stress that these specific functional forms are illustrative; the S3TS framework can accommodate any pair $(\priceFormula, \responseFormula)$ that admits black-box simulation. More elaborate response models, such as the data-driven ones considered in~\cite{pavirani2025predicting,madahi2024distributional}, can be plugged in without modification of the algorithm.
In this formulation, a quadratic cost exponent from \cref{eq:deterministic_price_publication_problem,eq:stochastic_price_publication_problem} is also used ($\costExponent \doteq 2$), more aggressively penalizing large publication errors than the linear case. 

The resulting problem combines three sources of analytical complexity: 
\begin{enumerate*}[(i)]
    \item the discontinuous price ladder $\priceFormula$, which would require introducing one binary indicator per step at every leaf of the scenario tree (\ie $\mathcal{O}(|\scenarioTreeLeaves|)$ binaries for each of the seventeen levels in \cref{fig:price_function}); 
    \item the non-convex quadratic-saturated response $\responseShape$, which couples $\price$ and $\SI$ through equality constraints that are neither convex nor concave, and which interacts multiplicatively with the time-dependent factor $\timeFactor(\timestep)$; and 
    \item the quadratic publication cost.
\end{enumerate*}
Together, these yield a non-convex Mixed-Integer Quadratically-Constrained Program. While such problems can in principle be encoded in modern solvers (\eg Gurobi~\cite{gurobi}), obtaining their global solution within a receding-horizon time budget is generally infeasible. By contrast, S3TS evaluates $\priceFormula$ and $\responseFormula$ only as black-box simulators along sampled trajectories, and remains therefore unaffected by their analytical complexity.
\section{Methodology}
\label{sec:methodology}

\subsection{Scenario Tree Construction}
\label{sec:scenario_tree}
 
To solve multi-stage stochastic optimization problems such as the one in \cref{eq:stochastic_price_publication_problem}, we assume the existence of a scenario tree $\scenarioTree$ that approximates the underlying random distribution influencing the problem dynamics. Scenario trees are required as inputs by both the stochastic MPC and the S3TS algorithm. We construct them in three sequential phases\,---\,scenario sampling, scenario reduction, and tree assembly\,---\,followed by a node-value refinement step, all described below. The full pipeline is illustrated in phases~1--3 of \cref{fig:s3ts_loop}.

\subsubsection{Scenario Sampling}
\label{sec:scenario_sampling}
We first sample a finite number of scenarios (\ie trajectories obtained through sequential realizations of the random variables governing the system) $\{ \scenario_\scenarioIndex \in \mathbb{R}^{\scenarioDimension \times \scenarioLength} \}_{\scenarioIndex=1}^{\numScenarioOriginal}$ from the underlying stochastic distribution, where $\numScenarioOriginal\in\mathbb{N}^+$ is the number of sampled trajectories, $\scenarioLength \in \mathbb{N}^+$ is the prediction horizon, and $\scenarioDimension \in \mathbb{N}^+$ is the dimension of each per-step realization ($\scenarioDimension \doteq 1$ in the problem considered in \cref{eq:stochastic_price_publication_problem}, as the scenarios only entail the fluctuations of future SI scalar values). Each scenario $\scenario_\scenarioIndex$ has an associated probability $\scenarioProbability_{\scenario_\scenarioIndex}$ such that $\sum_{\scenarioIndex=1}^{\numScenarioOriginal}{\scenarioProbability_{\scenario_\scenarioIndex}} = 1$. With uniform sampling, ${\scenarioProbability_{\scenario_\scenarioIndex} = \frac{1}{\numScenarioOriginal}} \; , \; \forall \scenarioIndex=1, 2, \dots, \numScenarioOriginal$.
 
In our experiments, scenarios are drawn directly from the same stochastic process used to generate the test sequences. This is a deliberate simplification, intended to keep the focus on the control algorithms rather than on the quality of the distribution approximation. In a deployment setting, scenarios would instead be produced by a probabilistic forecasting model fitted on historical data\,---\,for instance, the technique described in~\cite{pinson2009probabilistic}, or any model capable of generating samples consistent with the conditional predictive distribution. The framework used by S3TS imposes no specific assumption on how scenarios are produced and is therefore agnostic to the choice of the underlying forecasting model.
 
\subsubsection{Scenario Reduction}
\label{sec:scenario_reduction}
Sampling a large number $\numScenarioOriginal$ of scenarios is desirable to faithfully approximate the underlying distribution. However, feeding all of them into a multi-stage optimization problem is computationally prohibitive: the number of decision variables in the resulting optimization grows linearly with the number of scenarios retained, and so does the size of the search tree explored by S3TS. To mitigate this, we reduce the original scenario set to a smaller representative subset of $\numScenarioReduced \leq \numScenarioOriginal$ scenarios, using the algorithms described in~\cite{heitsch2003scenario,growe2003scenario}. These methods iteratively remove the least informative scenarios according to a Wasserstein-based distance metric on the trajectory space, and redistribute the probability mass of each removed scenario to its closest retained neighbor. The result is a reduced set $\{ \scenario_\scenarioIndex\}_{\scenarioIndex=1}^{\numScenarioReduced}$ with updated probabilities still satisfying $\sum_{\scenarioIndex=1}^{\numScenarioReduced}{\scenarioProbability_{\scenario_\scenarioIndex}} = 1$, and that approximates the original empirical distribution while remaining computationally tractable. The choice of $\numScenarioReduced$ is a trade-off: a larger value provides a more faithful approximation but potentially yields a larger tree and therefore a higher computational cost in the downstream optimization.
 
\subsubsection{Tree Assembly}
\label{sec:tree_assembly}
The reduced scenario set is then transformed into a scenario tree $\scenarioTree$. Following the principles laid out in~\cite{heitsch2009scenario}, we adopt a path-consistent agglomerative procedure based on stage-wise clustering on path prefixes, summarized in \cref{alg:tree_construction}.
 
The procedure starts from a single root node containing all reduced scenarios, and progressively partitions them stage by stage. At each stage $\timestep = 0, 1, \dots, \scenarioLength-1$, every existing tree node\,---\,corresponding to a subset $\scenarioSubset_\scenarioTreeNode$ of scenarios\,---\,is split into at most $\branchingFactor_\timestep$ children by clustering its scenarios according to their realizations up to (and including) time $\timestep$.
 
In practice, we use the K-Means algorithm~\cite{lloyd1982least} to perform clustering at each node, with the requested $\branchingFactor_\timestep$ as the number of clusters. The branching schedule $\{\branchingFactor_\timestep\}_{\timestep=0}^{\scenarioLength-1}$ is a hyperparameter of the construction. In our experiments, we use $\branchingFactor_0 = 1$, $\branchingFactor_1 = 3$, $\branchingFactor_2 = 3$, and $\branchingFactor_\timestep = 1$ for $\timestep \geq 3$, concentrating the branching in the early stages where the impact on the immediate decision is largest. This is a common heuristic in stochastic programming to control the otherwise exponential growth of the number of leaves while preserving the most relevant uncertainty splits. A node with $\branchingFactor_\timestep = 1$ effectively continues each scenario subset along a single branch, modeling the simplification that, once early-stage uncertainty has been resolved, residual uncertainty in later stages is approximated as a single conditional trajectory.
 
\begin{algorithm}
\caption{Scenario Tree Construction.}
\label{alg:tree_construction}
\begin{algorithmic}
\Require Reduced scenarios $\{ \scenario_\scenarioIndex \}_{\scenarioIndex=1}^{\numScenarioReduced}$, branching schedule $\{\branchingFactor_\timestep\}_{\timestep=0}^{\scenarioLength-1}$
\Ensure Scenario tree $\scenarioTree$
\State Initialise root node $\scenarioTreeNode_0$ with $\scenarioSubset_{\scenarioTreeNode_0} \gets \{1, \dots, \numScenarioReduced\}$
\State $\mathcal{N}_\text{current} \gets \{\scenarioTreeNode_0\}$
\For{$\timestep = 0, 1, \dots, \scenarioLength-1$}
    \State $\mathcal{N}_\text{next} \gets \emptyset$
    \For{each node $\scenarioTreeNode \in \mathcal{N}_\text{current}$}
        \State $k \gets \min\!\bigl(\branchingFactor_\timestep,\ |\scenarioSubset_\scenarioTreeNode|,\ \bigl|\{\scenario_\scenarioIndex[\timestep]\!:\,\scenarioIndex \in \scenarioSubset_\scenarioTreeNode\}\bigr|\bigr)$
        \State $X \gets \{\scenario_\scenarioIndex[0{:}\timestep+1] : \scenarioIndex \in \scenarioSubset_\scenarioTreeNode\}$
        \State $\{C_1, \dots, C_k\} \gets \Call{K-Means}{X, k}$
        \For{each cluster $C_j$}
            \State Create child node $\scenarioTreeNode'$ with $\scenarioSubset_{\scenarioTreeNode'} \gets C_j$
            \State Attach $\scenarioTreeNode'$ as child of $\scenarioTreeNode$ in $\scenarioTree$
            \State $\mathcal{N}_\text{next} \gets \mathcal{N}_\text{next} \cup \{\scenarioTreeNode'\}$
        \EndFor
    \EndFor
    \State $\mathcal{N}_\text{current} \gets \mathcal{N}_\text{next}$
\EndFor
\State Set $\scenarioProbability_\scenarioTreeNode \gets \sum_{\scenario \in \scenarioSubset_\scenarioTreeNode} \scenarioProbability_\scenario$ for every $\scenarioTreeNode \in \scenarioTree$
\State \Return $\scenarioTree$
\end{algorithmic}
\end{algorithm}
 
\subsubsection{Node-Value Refinement}
\label{sec:node_refinement}
Once the tree topology is fixed, each node $\scenarioTreeNode$ is assigned a representative realization $\bar{\scenario}_{\scenarioTreeNode} \in \mathbb{R}^{\scenarioDimension}$, computed as the probability-weighted conditional mean of the scenarios assigned to that node (indicated with $\scenarioSubset_\scenarioTreeNode$), evaluated at the node's time index $\timestep_\scenarioTreeNode$:
\begin{equation}
\label{eq:node_value_refinement}
    \bar{\scenario}_{\scenarioTreeNode} \doteq \frac{\sum_{\scenarioIndex \in \scenarioSubset_\scenarioTreeNode} \scenarioProbability_{\scenario_\scenarioIndex}\, \scenario_\scenarioIndex[\timestep_\scenarioTreeNode]}{\sum_{\scenarioIndex \in \scenarioSubset_\scenarioTreeNode} \scenarioProbability_{\scenario_\scenarioIndex}}.
\end{equation}
The trajectories along each root-to-leaf path are then redefined as the sequence of these conditional means, replacing the original raw scenarios. 
The refinement is what makes the node-level realization $\bar{\scenario}_{\scenarioTreeNode}$ well-defined as a property of the node alone. 
After the refinement, all scenarios sharing a path prefix in the tree share the same realizations along that prefix, which is the non-anticipativity condition required for the transition and reward functions as further discussed in \cref{sec:mdp_definition}. 
The choice of the conditional mean as the representative is, in turn, the natural one under a quadratic distance.
 
\subsubsection{Receding-Horizon Tree Construction}
\label{sec:receding_horizon_trees}
Both the stochastic MPC and the S3TS algorithm operate in a receding-horizon fashion: at every timestep $\timestep$ within the settlement period, a new tree is constructed using only the residual horizon $\scenarioLength - \timestep$ (forecasting the SI beyond the end of the settlement period does not help the optimization, as the objective function optimizes over each settlement period independently). For a settlement period of length $\settlementLength$, this yields $\settlementLength$ separate scenario trees per period, each with a progressively shorter horizon as time advances and uncertainty about the past is resolved by observation. Sampling, reduction, and assembly are repeated independently at each step, conditioned on the latest observed state.
 
For consistency, we use exactly the same scenario trees for stochastic MPC and S3TS within a given experiment, and we share them across techniques whenever applicable. This guarantees that any performance gap between the two methods is attributable to the optimization procedure itself rather than to differences in how uncertainty is represented.
 
\subsubsection{Notation Summary}
\label{sec:scenario_tree_notation}
For the description of the S3TS algorithm in the remainder of the paper, we collect the notation introduced above. We indicate the reduced scenario set used to build the tree as $\scenarioSet \doteq \{ \scenario_\scenarioIndex \}_{\scenarioIndex=1}^{\numScenarioReduced}$. Consistently with~\cref{eq:stochastic_price_publication_problem}, we denote the nodes of the tree by $\scenarioTreeNode \in \scenarioTree$. Each node has an associated scenario subset $\scenarioSubset_{\scenarioTreeNode} \in \mathcal{P}(\scenarioSet)$, a node-level realization $\bar{\scenario}_{\scenarioTreeNode}$ obtained from~\cref{eq:node_value_refinement}, and a probability value $\scenarioProbability_{\scenarioTreeNode} \doteq \sum_{\scenario \in \scenarioSubset_\scenarioTreeNode}{\scenarioProbability_{\scenario}}$. Finally, we describe the dynamics of the parental relations between nodes through a \textit{\textualScenarioNode map} $\splitterFunction : \scenarioTree \rightarrow \mathcal{P}(\scenarioSet)$, which describes how the scenarios' uncertainty evolves through the tree by indicating, for each node, the partition induced on its scenario subset by its children.

\subsection{MDP definition}
\label{sec:mdp_definition}
Before describing the S3TS algorithm, we need to formalize the mathematical framework it works with, \ie a Markovian Decision Process (MDP)~\cite{busoniu2017reinforcement}.
An MDP is a sequential decision process model typically used with Reinforcement Learning (RL) algorithms, including the MCTS algorithm we expanded to build S3TS. 
Fixing a scenario $\scenario$ (\ie SI trajectory), we obtain a \textit{deterministic}\footnote{By fixing a scenario as in the definition in \cref{sec:scenario_tree}, the process becomes deterministic as there is no uncertainty involved in the future values.} MDP defined as the tuple $\bigl( \stateSpace, \actionSpace, \transitionFunction_{\scenario}, \rewardFunction_{\scenario} \bigr)$, where
\begin{itemize}
    \item $\stateSpace$ is the \textit{State Space},
    \item $\actionSpace$ is the \textit{Action Space},
    \item $\transitionFunction_{\scenario} : \stateSpace \times \actionSpace \rightarrow \stateSpace$ is the \textit{Transition Function},
    \item $\rewardFunction_{\scenario} : \stateSpace \times \actionSpace \rightarrow \mathbb{R}$ is the \textit{Reward function}.
\end{itemize}
By considering a set of scenarios $\scenarioSet$, we obtain an approximation of the underlying stochastic process in the form of a set of deterministic MDPs $\{ \bigl( \stateSpace, \actionSpace, \transitionFunction_{\scenario}, \rewardFunction_{\scenario} \bigr) \; ; \; \forall \scenario \in \scenarioSet \}$. Given a discount factor $\gamma \in [0, 1]$, the corresponding objective would then become to find the optimal policy $\pi^* : \stateSpace \rightarrow \actionSpace$ such that:

\begin{equation}
    \pi^*(\MDPstate) \in \arg \max_{\action \in \actionSpace}{Q^*(\MDPstate, \action)} \; ; \; \forall \MDPstate \in \stateSpace \; .
\end{equation}
where:

\begin{align}
    Q^*(\MDPstate, \action) & \doteq \max_\pi{Q^\pi(\MDPstate, \action)}, \\
    Q^\pi(\MDPstate, \action) & \doteq \mathbb{E}_{\scenario \sim \scenarioSet} \Bigl[ \rewardFunction_\scenario(\MDPstate, \action) + \gamma R^{\pi}\Bigl( \transitionFunction_\scenario(\MDPstate, \action) \Bigr) \Bigr] \\
    & = \sum_{\scenario \in \scenarioSet}^{}{\scenarioProbability_\scenario \Bigl[ \rewardFunction_\scenario (\MDPstate, \action ) + \gamma R_\scenario^{\pi} \Bigl( \transitionFunction_\scenario(\MDPstate, \action) \Bigr) \Bigr]}  \nonumber, \\
    R^{\pi}(\MDPstate_0) & \doteq \lim_{T\rightarrow +\infty}\mathbb{E}_{\scenario\sim\scenarioSet}\left[ {\sum_{t=0}^{T}{\gamma^t\rewardFunction_\scenario\Bigl(\MDPstate_\timestep, \pi(\MDPstate_\timestep)\Bigr)}} \right] \\
    & = \lim_{T\rightarrow +\infty} \sum_{\scenario \in \scenarioSet}^{}{\left[ \scenarioProbability_\scenario {\sum_{\timestep=0}^{T}{\gamma^t \rewardFunction_\scenario \Bigl( \MDPstate_\timestep, \pi(\MDPstate_\timestep) \Bigr)}} \right]} \nonumber , \\
    R_\scenario^{\pi}(\MDPstate_0)  &\doteq \lim_{T\rightarrow +\infty}\sum_{\timestep=0}^{T}{\gamma^t\rewardFunction_{\scenario}(\MDPstate_\timestep, \pi(\MDPstate_\timestep))} \\ 
     & \text{with } \; \MDPstate_{\timestep+1} = \transitionFunction_{\scenario}(\MDPstate_\timestep, \action) , \, \forall \timestep \in \mathbb{N} \nonumber
\end{align}

Notice that the optimal policy $\pi^*$ as we just defined is not equivalent to the optimal policy of the stochastic process we wish to solve. Instead, $\pi^*$ is only the optimal policy for the aggregated set of MDPs $\{ \bigl( \stateSpace, \actionSpace, \transitionFunction_{\scenario}, \rewardFunction_{\scenario} \bigr) \; ; \; \forall \scenario \in \scenarioSet \}$, which is on its own an approximation of the actual underlying process we wish to solve. The quality of the policy $\pi^*$ will then also depend on the quality of such an approximation; in other words, it will depend on how accurate the scenario tree used is in relation to the real random distribution steering the values of the stochastic variables involved. 

As discussed in \cref{sec:scenario_tree}, $\scenarioSet$ has been sampled and reorganized to build a scenario tree $\scenarioTree$. To ensure that the just-defined optimization problem is consistent with the non-anticipativity property given by the scenario tree, we then impose the following conditions. Given a node in the scenario tree $\scenarioTreeNode \in \scenarioTree$, and a pair of state and action $\MDPstate \in \stateSpace$, $\action \in \actionSpace$, we have:
\begin{itemize}
    \item $\transitionFunction_\scenario (\MDPstate, \action) = \transitionFunction_{\scenario'} (\MDPstate, \action) \; ; \; \forall \scenario, \scenario' \in \scenarioSubset_\scenarioTreeNode$
    \item $\rewardFunction_\scenario (\MDPstate, \action) = \rewardFunction_{\scenario'} (\MDPstate, \action) \; ; \; \forall \scenario, \scenario' \in \scenarioSubset_\scenarioTreeNode$
\end{itemize}
For simplicity, we will indicate the corresponding functions with $\transitionFunction_{\scenarioTreeNode}$ and $\rewardFunction_{\scenarioTreeNode}$.

In the price publication problem defined in \cref{sec:problem_description}, the states of the MDP would indicate the general condition of the system (\eg SI, time), the actions would be the published price ($\price$), the transition function would be represented by the functions $\SIProgressionFormula(\SI), \responseFormula(\price)$, and the reward function would entail the objective functions defined in \cref{eq:deterministic_price_publication_problem,eq:stochastic_price_publication_problem}.
Last, a definition of intermediate $\rewardFunction$ is required for the tree-based methods. 
In fact, the optimization cost in \cref{eq:deterministic_price_publication_problem,eq:stochastic_price_publication_problem} is defined at the end of the settlement period, once the final price $\hat{\price}_\settlementLength$ is known. 
Tree search methods, however, backpropagate values through the search tree at every step, and therefore require an intermediate reward $\rewardFunction_{\scenarioTreeNode}$ at each node. 
We define this intermediate reward at timestep $\timestep$ in the settlement period as the negative mean deviation between the prices published so far in the period and the running estimate of the final price obtained from the current cumulative average SI:
\begin{equation}
\label{eq:intermediate_reward}
    \rewardFunction_\timestep \;\doteq\; -\,
    \frac{1}{\timestep}\sum_{\tau=0}^{\timestep-1}
    \bigl|\price_\tau - \priceFormula(\averageSI_\timestep)\bigr|^{\costExponent}.
\end{equation}
At $\timestep = \settlementLength$ the rolling estimate $\priceFormula(\averageSI_\settlementLength)$ coincides with $\hat{\price}_\settlementLength$ and the sum of per-step rewards recovers the total period cost in \cref{eq:deterministic_price_publication_problem}. 
This shaping is needed only for the tree search methods, since the mathematical-programming techniques operate directly on the end-of-period cost and do not require an intermediate signal.

\subsection{Monte Carlo Tree Search}
The MCTS algorithm models the MDP using a search tree structure, mapping the states to nodes and the actions to edges. The algorithm is a sequential repetition of four phases, roughly described as follows:
\begin{enumerate}
    \item \textbf{Selection:} Starting from a root node $\MDPstate_0$, iteratively traverse the tree by reaching a child node (\ie a system state) ${\MDPstate_\depth \; ; \; \depth = 1, \dots, \leafNode}$ by selecting an edge (action) ${\action_\depth \; ; \; \depth = 0, \dots, \leafNode-1}$ until a leaf node $\MDPstate_\leafNode$ is reached. Each selection is made by balancing exploitation (selecting the actions that proved to be the most optimal in the previous iterations) and exploration (selecting the actions that have not been selected enough before, with the goal to find more profitable trajectories compared to the ones explored so far).
    \item \textbf{Expansion:} Expand the tree by adding new children from the selected leaf node $\MDPstate_\leafNode$, \ie roll out possible actions from that state.
    \item \textbf{Simulation:} Evaluate the node value by performing Monte Carlo simulations. 
    \item \textbf{Backpropagation:} Propagate the information acquired (\ie the rewards obtained in the trajectory) back to the root node. This information will update the value estimation of each node in the selected trajectory. These value estimations will then be used in the selection phase to indicate how promising actions are in the tree.
\end{enumerate}
The execution of the four phases constitutes an iteration (or simulation). The more iterations are executed, the more accurate the tree search will become, improving the control efficacy of the policy.
Notice that the nodes and edges of the search tree are identified solely by a depth value $\depth \in \mathbb{N}$. In general, a tree may contain multiple nodes and edges at the same depth. However, within the framework of the MCTS algorithm, such a finer distinction is unnecessary. During a single iteration, exactly one node and one edge are selected at each depth level. Therefore, referring to nodes and edges only by their depth is sufficient and does not introduce ambiguity for the purposes of the algorithm.

\begin{algorithm}[H]
\caption{S3TS Algorithm.}
\label{alg:stochastic_ts}
\begin{algorithmic}
\Procedure{Simulation\_Iter}{$\MDPstate_\depth^\scenarioSubset$}
    \If{$\MDPstate_\depth^\scenarioSubset$ is not a leaf}
        \State Select action $\action_\depth^\scenarioSubset$ \Comment{Based on \cref{eq:selection_formula}}
        \State Get $\scenarioTreeNode_\depth^\scenarioSubset$ from $\MDPstate_\depth^\scenarioSubset, \action_\depth^\scenarioSubset$ \Comment{Corresponding scenario tree node}
        \State \texttt{v\_list} $\gets [\;]$
        \For{$\scenarioSubset'$ in $\splitterFunction( \scenarioTreeNode_\depth^\scenarioSubset )$}
            \State $\MDPstate_{\depth+1}^{\scenarioSubset'} \gets \transitionFunction_{\scenarioTreeNode_\depth^{\scenarioSubset'}} (\MDPstate_\depth^\scenarioSubset,\action_\depth^\scenarioSubset)$  \Comment{Transition function}
            \State $\reward_{\depth}^{\scenarioSubset'} \gets \rewardFunction_{\scenarioTreeNode_\depth^{\scenarioSubset'}}(\MDPstate_\depth^\scenarioSubset,\action_\depth^\scenarioSubset)$ \Comment{Reward function}
            \State Append $\Bigl($\Call{Simulation\_Iter}{$\MDPstate_{\depth+1}^{\scenarioSubset'}$}$+\reward_{\depth}^{\scenarioSubset'},\; \scenarioSubset'\Bigr)$ to \texttt{v\_list}
        \EndFor
        \State \texttt{v} $\gets$ \Call{Aggregate}{\texttt{v\_list}}
        \State Update $\overline{Q}_\text{tree}\Bigl(\MDPstate_\depth^\scenarioSubset, \action_\depth^\scenarioSubset\Bigr)$ \Comment{Based on \cref{eq:mcts_q_update}}
    \Else
        \State \Call{Expand}{$\MDPstate_\depth^\scenarioSubset$}
        \State \texttt{v} $\gets 0$
    \EndIf
    \State \Return \texttt{v}
\EndProcedure
\State

\Procedure{Aggregate}{\texttt{v\_list}}
    \State \texttt{v\_total} $\gets 0$
    \For{$($\texttt{v}$, \,\scenarioSubset')$ in \texttt{v\_list}}
        \State\texttt{v\_total} $\gets$ \texttt{v\_total} $+$ \texttt{v} $\mathbb{P}(\scenarioSubset')$
    \EndFor
    \State \Return \texttt{v\_total}
\EndProcedure
\State

\Procedure{Expand} {$\MDPstate_\depth^\scenarioSubset$}
    \For{$\action$ in $\actionSpace$}
        \State Get $\scenarioTreeNode_\depth^\scenarioSubset$ from $\MDPstate_\depth^\scenarioSubset, \action$ \Comment{Corresponding scenario tree node}
        \For{$\scenarioSubset'$ in $\splitterFunction( \scenarioTreeNode_\depth^\scenarioSubset )$}
            \State Add $\MDPstate_{\depth+1}^{\scenarioSubset'}$ into the tree
        \EndFor
    \EndFor
\EndProcedure
\State

\Procedure{Stochastic\_Tree\_Search} {$\MDPstate_0^\scenarioSet$}
    \While{computational budget is available}
        \State \Call{Simulation\_Iter}{$\MDPstate_0^\scenarioSet$}
    \EndWhile
    \State \Return $\arg \max_{\action\in\actionSpace}{\left[Q_{\text{tree}}\Bigl(\MDPstate_0^\scenarioSet,\action\Bigr)\right]}$ \Comment{Best action}
\EndProcedure
\end{algorithmic}
\end{algorithm}

In our experiments, the Selection phase uses the following equation based on the UCT formula~\cite{kocsis2006bandit}:
\begin{equation}
\label{eq:selection_formula} 
    \action_\depth = \arg \max_{\action \in \mathcal{\actionSpace}^\depth} \biggl\{ \overline{Q}_\text{tree}(\MDPstate_\depth, \action) + \alpha \frac{\sqrt{N(\MDPstate_\depth)}}{1+N(\MDPstate_\depth, \action)} \biggl\} \; ,
\end{equation}
where $N(\MDPstate) \in \mathbb{N}$ is the number of visits received by node $\MDPstate$ (\ie the number of time the nodes has been selected before); $N(\MDPstate, \action) \in \mathbb{N}$ is the number of times action $\action$ has been selected from state $s$; $\alpha \in \mathbb{R}^+$ is a hyperparameter used to balance exploration and exploitation; and $\overline{Q}_{\text{tree}}(s, a)$ is the normalized version of a state-action value function approximation that is tuned in the backpropagation phase as follows:
\begin{align}
\label{eq:mcts_q_update} 
    Q_\text{tree}(\MDPstate_\depth, \action_\depth) = &
    \frac{N(\MDPstate_\depth, \action_\depth)\> Q_\text{tree}(\MDPstate_\depth, \action_\depth) + G^{k+1}}{N(\MDPstate_\depth, \action_\depth) + 1} \;, \\
    \nonumber & \forall \depth = \leafNode-1, \dots, 0
\end{align}
where $Q_\text{tree}(\MDPstate_\depth, \action_\depth)$ is initialized as $\rewardFunction(\MDPstate_\depth, \action_\depth)$ and $G^k$ is the discounted accumulated reward in the tree branch defined as:
\begin{align}
    G^k = 
    \begin{cases}
      \rewardFunction(\MDPstate_{\leafNode-1}, \action_{\leafNode-1}) & \text{if }  k = \ell\\
      \rewardFunction(\MDPstate_{\depth-1}, \action_{\depth-1}) + \gamma \> G^{k+1} & \text{if }  k = \ell-1, \dots, 1
    \end{cases} \;;
\end{align}
with $\gamma \in [0, 1]$ the discount factor. The normalized value $\overline{Q}_\text{tree}(\MDPstate, \action)$ is obtained using min/max values in the tree:
\begin{equation}
        \overline{Q}_\text{tree}(\MDPstate, \action) = \frac{Q(\MDPstate, \action) - Q^\text{min}}{Q^\text{max} - Q^\text{min}} \in [0, 1] \;\; ,
    \end{equation}
where 
\begin{equation}
    \begin{cases}
        Q^\text{min} \doteq \min_{\MDPstate', \action' \in \text{Tree}}{Q_\text{tree}(\MDPstate', \action')} \\
        Q^\text{max} \doteq \max_{\MDPstate', \action' \in \text{Tree}}{Q_\text{tree}(\MDPstate', \action')}
    \end{cases} \;.
\end{equation}

\begin{figure*}
    \centering
    \includegraphics[width=1\textwidth]{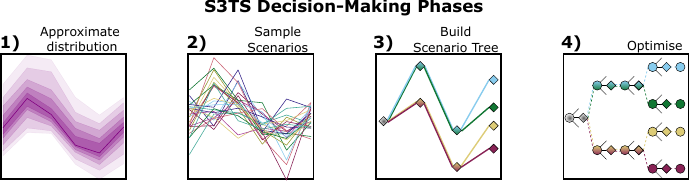}
    \caption{Four phases repeated at each timestep to use the S3TS algorithm. First, the random distribution underneath the stochastic process gets approximated (phase 1). Then, using the approximated distribution, a certain number of scenarios are sampled from it (phase 2). The scenarios get aggregated in a scenario tree (phase 3) that is then optimized using the S3TS algorithm (phase 4), obtaining the optimal action from it. Notice that the structure is very similar to the generic stochastic MPC one, where the first three phases are equivalent, and phase 4 can be changed with other optimization algorithms (\eg mathematical optimization).}
    \label{fig:s3ts_loop}
\end{figure*}

\subsection{Stochastic Scenario-Structured Tree Search (S3TS)}
\label{sec:S3TS_methodology}

The standard MCTS algorithm was developed with the goal of working with deterministic MDPs. That is because it is expected that, given a certain state (node), an action (edge) would deterministically and uniquely lead to a defined next state. To allow the algorithm to handle uncertainties in the problem, we extended the MCTS algorithm described above by integrating scenario trees in the tree search. To do so, we introduce the distinction between state nodes (\ie the nodes mapped to a state of the control problem (MDP), as used in standard MCTS) and \textit{\textualScenarioNode nodes} (\ie nodes that are mapped with the nodes of a scenario tree), and action edges (\ie the edges mapped to an action of the control problem, as used in standard MCTS) with \textit{\textualScenarioNode edges} (\ie the edges that are mapped to the edges of a scenario tree). Roughly speaking, we naturally integrate a scenario tree in a search tree by adding its corresponding \textualScenarioNode nodes and edges between the state nodes. In particular, each action edge will now go from a state node (the state in which the control action takes place) to a \textualScenarioNode node mapped to the corresponding node in the scenario tree. 
Then, the \textualScenarioNode node will solve the uncertainty dynamics of the scenario tree by having multiple \textualScenarioNode edges ending at a state node, each corresponding to a specific child of the scenario tree node. Note that, differently from the deterministic version, the depth $\depth$ is now not sufficient to indicate each selected node and action. Instead, it is now necessary to also indicate the corresponding subset of scenarios $\scenarioSubset \in \mathcal{P}(\scenarioSet)$ that the node is associated with (based on the scenario tree structure). For this reason, state nodes and action edges are now indicated with the notation $\MDPstate_\depth^\scenarioSubset$, $\action_\depth^\scenarioSubset$, respectively. 
Analogously, we indicate with $\scenarioTreeNode_\depth^\scenarioSubset$ the corresponding node in the scenario tree and with $\reward_\depth^\scenarioSubset$ the corresponding reward obtained. 

\begin{figure}
    \centering
    \includegraphics[width=0.5\textwidth]{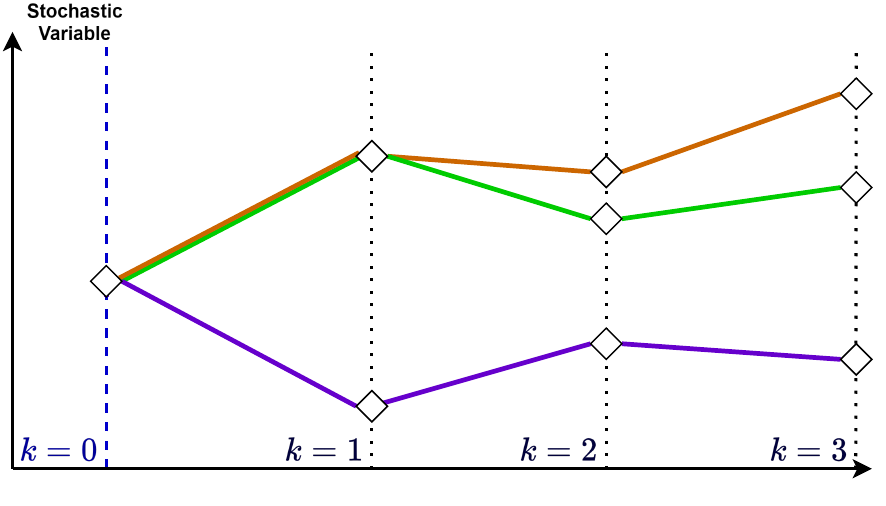}
    \caption{Example of a scenario tree with horizon 3}
    \label{fig:scenario_tree}
\end{figure}

\begin{figure}
    \centering
    \includegraphics[width=0.5\textwidth]{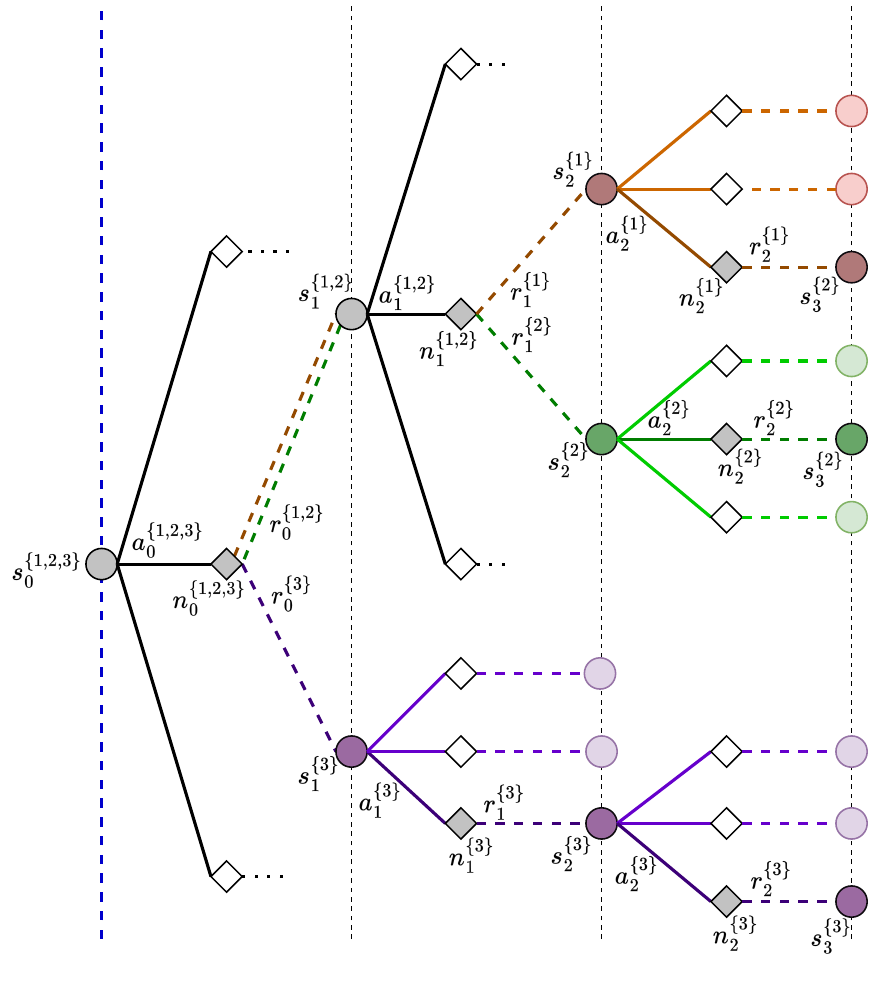}
    \caption{Notation of the S3TS framework; using $\{1,2,3\}$ as the scenario space with 1 representing the orange scenario of \cref{fig:scenario_tree}, 2 the green one, and 3 the purple one. Darker nodes indicate the selected trajectories in the current tree search iteration.}
    \label{fig:stochastic_tree_search}
\end{figure}

At a high level, the algorithm follows the same four-phase loop as standard MCTS\,---\,selection, expansion, simulation, and backpropagation\,---\,but treats stochastic transitions explicitly.
Starting from the root state node, the algorithm uses the UCT criterion of \cref{eq:selection_formula} to select an action; this action leads deterministically to a \textualScenarioNode node mapped onto the
corresponding node of the scenario tree. 
From there, the \textualScenarioNode node resolves the stochastic transition by branching into all of its scenario children, and the algorithm recurses into each child state node. The values returned by the recursive calls are aggregated into a single probability-weighted estimate, which is backpropagated up the tree to update the $Q_\text{tree}$ value of the selected action via \cref{eq:mcts_q_update}. 
When the recursion reaches a state node that has not yet been expanded, the algorithm grows the tree at that point. 
For each available action, it adds a \textualScenarioNode node and, for each child of that \textualScenarioNode node, a new state node. 
The current simulation then terminates without recursing further; subsequent simulations that reach this point will be able to descend through the newly added children.
This loop is repeated until the computational budget available for each action is exhausted, at which point the action with the highest $Q_\text{tree}$ value at the root is returned.
An example of a search tree integrated with a scenario tree is shown in \cref{fig:stochastic_tree_search} corresponding to the scenario tree in \cref{fig:scenario_tree}. The full algorithm is described in \cref{alg:stochastic_ts}.\footnote{The implementation code is available at \\ {https://github.com/loud-mime/S3TS/tree/main}} The phases iterated upon to use the S3TS algorithm are depicted in \cref{fig:s3ts_loop}.


\section{Experimental Setup}
\label{sec:experimental_setup}
 
\subsection{Data Generation}
\label{sec:data_generation}
 
\begin{figure*}
    \centering
    \includegraphics[width=1\textwidth]{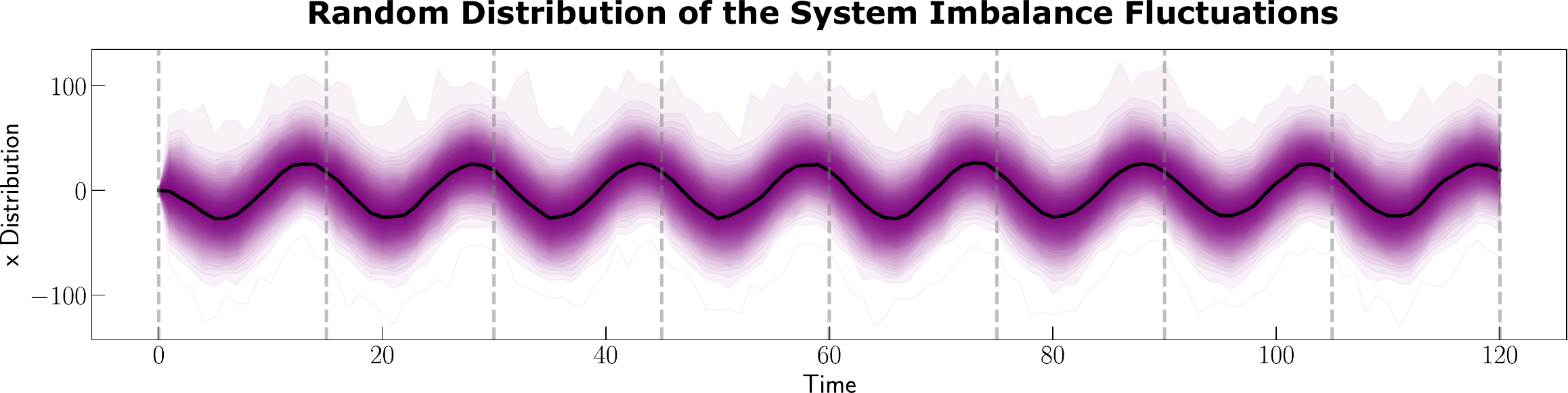}
    \caption{Visualization of the random distribution used to generate the SI fluctuations in our experiments. The distribution is approximated through Monte Carlo sampling, and different quantiles are visualized in the figure.}
    \label{fig:random_distribution}
\end{figure*}

To evaluate the algorithms on the problems described in \cref{eq:deterministic_price_publication_problem,eq:stochastic_price_publication_problem}, we considered a synthetic stochastic process $\SIFluctuationsDistribution$ governing the SI fluctuations of $\SIProgressionFormula$. We then generated a timeseries of datapoints sequentially sampled from $\SIProgressionFormula$ and divided it into $1000$ independent settlement periods of $\settlementLength \doteq 15$ timesteps of 1 minute each. Each period is approached independently by every algorithm to produce a sequence of published prices ($\price_\timestep \;;\; \forall \timestep = 0, 1, \dots, \settlementLength$). A graphical illustration of the underlying random distribution is shown in \cref{fig:random_distribution}, with probabilistic quantiles obtained through Monte Carlo sampling.
The process is defined as a first-order autoregressive stochastic system, following the structure used in \cref{eq:deterministic_price_publication_problem,eq:stochastic_price_publication_problem}. The seasonal progression component $\seasonalFunction(\timestep)$ follows a sinusoidal profile with period $15$ (corresponding to $\settlementLength$). The balancing factor is fixed to $\balancingFactor \doteq 0.5$. The stochastic fluctuations $\SIFluctuations$ are modeled as time-dependent normal noise using a cosine modulation. More precisely, the process can be written as
\begin{equation}
\SI_{\timestep+1} = 15\sin\!\left(\frac{2\pi \timestep}{15}\right) + (1-\balancingFactor)\SI_t + \SIFluctuations_\timestep,
\end{equation}
where 
\begin{equation}
\SIFluctuations_\timestep \sim \mathcal{N}\!\left(0,\sigma_t^2\right),
\qquad
\sigma_t = 20 + 5\cos\!\left(\frac{2\pi (\timestep \bmod 15)}{60}\right).
\end{equation}

To build the scenario trees used by the stochastic methods (Stochastic MPC and S3TS) when solving \cref{eq:stochastic_price_publication_problem}, at every timestep within each settlement period we drew $\numScenarioOriginal=10000$ SI trajectories from the underlying distribution conditional on the latest observed state, reduced them to $\numScenarioReduced=100$ representatives following \cref{sec:scenario_reduction}, and assembled the resulting scenario tree using the path-consistent procedure of \cref{alg:tree_construction}. For consistency, Stochastic MPC and S3TS share exactly the same scenario tree at every receding-horizon step within an experiment\,---\,this guarantees that any performance gap between the two methods is attributable to the optimization procedure itself rather than to differences in uncertainty representation. The deterministic-equivalent baselines are obtained by feeding the very same code path with degenerate trees containing a single trajectory (the element-wise median of the original scenario set). All techniques operate in a receding horizon, \ie in every timestep the forecasts are regenerated, then the problem is solved with the latest information, and only the upcoming action is applied before the iteration repeats.
 
\subsection{Baselines}
\label{sec:techniques}
 
The evaluation compares six techniques that span three pairings of interest: tree search against mathematical optimization, stochastic against deterministic uncertainty handling, and S3TS against a rule-based baseline. To avoid implementation artefacts confounding the comparison, every algorithm variant runs through the same code path and shares the same input scenario trees whenever applicable. The baselines based on mathematical solvers (\ie Perfect Knowledge and the Stochastic and Deterministic MPCs) are reported only in the linear regime experiment's results, since the non-linear regime is intractable for MILP.
\begin{itemize}
    \item \textbf{Rule-Based.} A naive heuristic that publishes the price obtained by applying $\priceFormula$ to the cumulative average SI observed so far in the current settlement period. This baseline carries no forecasting and no optimization, but it is usually applied in practice, and it serves to quantify the value of advanced planning.
    \item \textbf{Perfect Knowledge.} An MILP solution to \cref{eq:deterministic_price_publication_problem} computed under the assumption of perfect foresight\,---\,\ie with the realized SI trajectory of the period revealed in advance. This is a clairvoyant oracle, not a stochastic optimum, \ie it provides an unattainable upper bound on the theoretical best policy obtainable.
    \item \textbf{Stochastic MPC.} An MILP solution of \cref{eq:stochastic_price_publication_problem} given the input scenario tree. This is the optimum of the stochastic problem \emph{conditional on the input tree}, and is the most informative reference point for assessing S3TS: any gap between Stochastic MPC and S3TS measures how well S3TS approximates the optimum of the same multi-stage stochastic problem the two methods share.
    \item \textbf{Deterministic MPC.} The MILP solution of \cref{eq:deterministic_price_publication_problem} fed with the elementwise median of the sampled scenarios as a point forecast. This isolates the cost of replacing the scenario tree with a single representative trajectory, while keeping the optimization procedure (MILP) unchanged.
    \item \textbf{S3TS.} The algorithm proposed in this work, instantiated as in \cref{alg:stochastic_ts}.
    \item \textbf{MCTS.} The deterministic counterpart to S3TS, obtained by running \cref{alg:stochastic_ts} on a singleton scenario tree whose unique trajectory is the element-wise median of the original scenario set. This baseline shares its implementation, action space, expansion policy, UCT exploration constant, and value-update rule with S3TS; the only difference is the input scenario tree. As a consequence, any performance gap between this baseline and S3TS is attributable solely to the explicit handling of uncertainty in the planning framework, with all other algorithmic factors kept unchanged.
\end{itemize}
 
The tree search techniques (S3TS and MCTS) require a discrete action space; we adopt the same construction as in~\cite{pavirani2025predicting}, recomputed at every timestep based on the current SI estimate.
 
\subsection{Computational Budget Protocol}
\label{sec:budget_protocol}
 
A core methodological aim of our evaluation is to disentangle the contribution of S3TS from the contribution of the computational budget involved. Tree search methods can, in principle, be made arbitrarily accurate by running additional simulations. We therefore adopt a matched-budget protocol: every technique is given the same per-publication wall-clock budget, and we report results across a grid of budgets to expose how performance scales with available compute.
 
Concretely, we run each experiment at four budgets: $\budget \in \{1, 2, 4, 8\}$ seconds per published price. The budget is enforced as wall-clock time inside the inner solve loop: the loop checks elapsed time after every completed simulation and exits as soon as the budget is exceeded. The budget is applied uniformly to all techniques. The Rule-Based heuristic terminates in microseconds and is therefore always well below budget; the MILP solves (Perfect Knowledge, Stochastic MPC, Deterministic MPC) typically reach close-to-optimal results within a fraction of a second on the linear formulation, and consequently also finish below budget and are not affected by our variations of $\budget$.
 
\subsection{Validation Metrics}
\label{sec:metrics}
We evaluate each technique using two cost metrics that mirror the cost function of \cref{eq:deterministic_price_publication_problem,eq:stochastic_price_publication_problem}: the \emph{Mean Absolute Error} (MAE) corresponds to $\costExponent \doteq 1$, and the \emph{Mean Squared Error} (MSE) to $\costExponent \doteq 2$, both computed from the published prices ($\price_\timestep$) and the realized price $\hat{\price}_\settlementLength$ difference at the end of the settlement period. 
Since $\costExponent$ enters the objective each technique optimizes, we report three experiments: the linear regime under $\costExponent \doteq 1$ (MAE)\,---\,the only setting admitting a tractable MILP encoding thanks to the linear dynamics and therefore the only one we can benchmark against mathematical optimization (\ie the MPCs and Perfect Knowledge baselines); the non-linear regime under $\costExponent \doteq 1$ (MAE), which isolates the effect of the non-linear dynamics while keeping the cost exponent fixed; and the non-linear regime under $\costExponent \doteq 2$ (MSE), matching a quadratic-cost setting.
Each metric is averaged over the $1000$ independent settlement periods. Together with each mean, we report the corresponding inter-quartile (Q1--Q3) interval taken over the same $1000$ periods.

\section{Results}
\label{sec:results}

\subsection{Linear Model Results}
\label{sec:linear_results}
\Cref{tab:results_linear_q1} reports the MAE achieved by each technique on the linear regime with $\costExponent=1$ across the four budget settings. At $8$\,s, S3TS achieves $25.93$ MAE against Stochastic MPC's $22.81$, placing it within $13.7\%$ of the optimum of the multi-stage stochastic problem conditioned to the same input scenario tree; indicating S3TS to be a valuable control algorithm. 
The gap narrows with budget, from $15.4\%$ at $1$\,s to $13.7\%$, indicating that part of the distance to the optimum can also be attributed to the simulation budget rather than to a structural limitation of the algorithm (\eg discretization of the action space).

A separate and complementary observation concerns the two deterministic-equivalent techniques: Deterministic MPC and MCTS consume the same median scenario as input but differ in their solver, and the tree search method wins by a substantial margin (MAE of $27.65$ vs.~$32.67$ at $8$\,s, a $15.4\%$ advantage that is consistent across budgets). 
This is initially counter-intuitive: MPC solves the deterministic problem to optimality given the median forecast, whereas MCTS is a sample-based approximation working within a finite simulation budget\,---\,one might a priori expect MPC to do at least as well.
Our intuition is that the MPC's optimality is conditional on the forecast being correct across the entire settlement window, \ie it commits to a full sequence of decisions based on the whole median trajectory, propagating any forecast error all the way to the cost incurred at the end of the period. 
Because the SI random process accumulates the uncertainties later in the prediction horizon, the median trajectory is a particularly poor predictor of the late-period dynamics that ultimately determine $\hat{\price}_\settlementLength$. 
MCTS, by contrast, is implicitly horizon-limited: a limited simulation budget gives each rollout only finite depth, so the algorithm weights near-term consequences more heavily than far-future ones when selecting actions. 
This horizon bias is harmful in general, but protective when the long-horizon forecast is unreliable.

The same mechanism explains why the gap between S3TS and MCTS widens with budget rather than closing, from $4.6\%$ at $1$\,s to $6.6\%$ at $8$\,s, even though the two methods share their entire implementation and differ only in their input scenario tree. 
The MCTS trajectory in \cref{tab:results_linear_q1} is essentially flat across $1$--$4$\,s ($27.49$--$27.53$) and ticks up slightly at $8$\,s ($27.65$). 
A larger budget allows the algorithm to explore deeper into the simulation tree and incorporate more of the median trajectory into its value estimates\,---\,which, by the argument above, degrades the publication rather than improving it. 
The deterministic baseline thus shows a mild overfitting-to-forecast effect, asymptoting toward the failure mode of Deterministic MPC as compute time grows. 
S3TS does not exhibit this pattern because its scenario tree expresses the uncertainty in the late-period SI through multiple branches rather than committing to a single misleading trajectory: deeper search lets it weight those branches more accurately, not propagate a wrong point forecast further. 
The result is that uncertainty representation and compute budget interact constructively for S3TS and destructively for the deterministic baseline.

A sizeable gap separates all techniques from Perfect Knowledge ($15.34$ MAE), with $48.7\%$ remaining even for Stochastic MPC. 
We emphasize that Perfect Knowledge is not the optimum of the stochastic problem the techniques are solving: it is a clairvoyant oracle that observes the realized SI trajectory in advance, and as such represents an unattainable target. 
Its distance from Stochastic MPC quantifies the irreducible cost of stochasticity in this problem\,---\,even an optimal stochastic policy cannot, on average, close to within $48\%$ of clairvoyance under the SI fluctuations we considered. This is intrinsic to the random process, not to any specific algorithm, and is consistent with the empirical observations that publishing imbalance settlement prices remains a difficult problem even given strong forecasting tools.

\begin{figure*}
    \centering
    \includegraphics[width=1\textwidth]{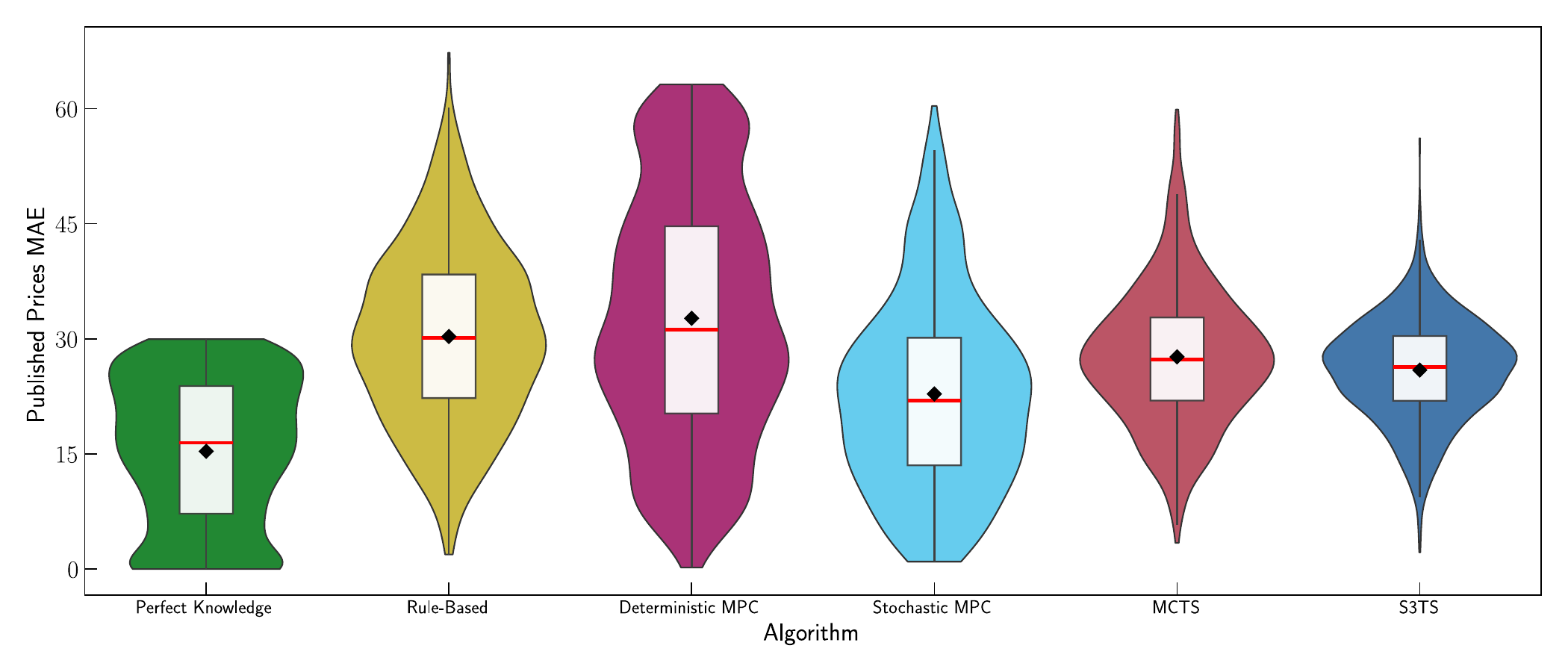}
    \caption{Distribution of the published price MAE vs. the actual one in the linear problem formulation. The red lines indicate the median values, while the black diamonds indicate the means.}
    \label{fig:linear_q1_violin}
\end{figure*}

\begin{figure*}
    \centering
    \includegraphics[width=1\textwidth]{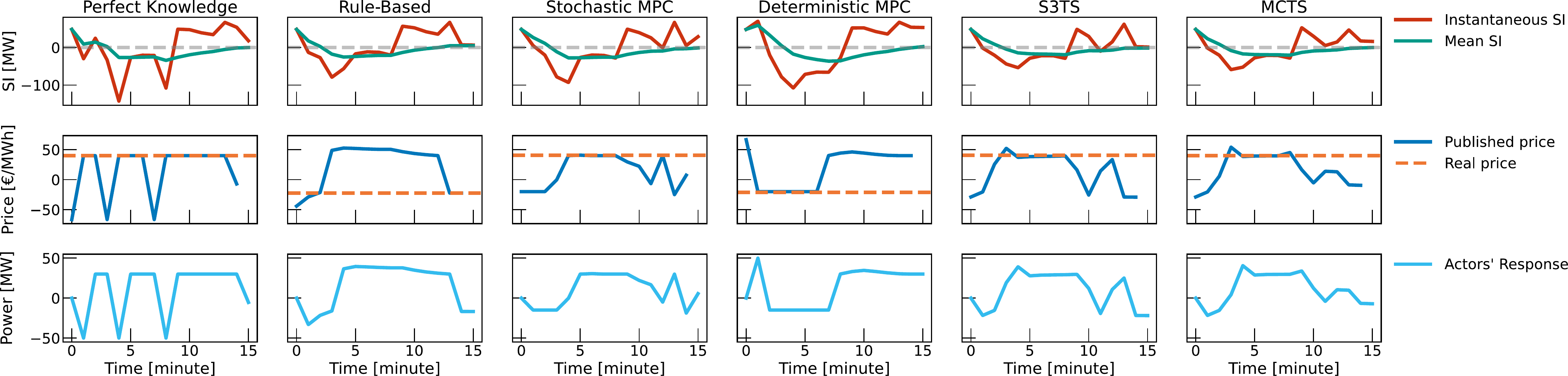}
    \caption{Example of price publication in the linear problem formulation using different techniques. The rolling average SI faces sign instability throughout the settlement period, adding significant uncertainty to the final price.}
    \label{fig:settlement_graph}
\end{figure*}

The distributional view in \cref{fig:linear_q1_violin} adds qualitative support to the table. 
Stochastic MPC and S3TS exhibit MAE distributions more concentrated than Deterministic MPC, which shows a longer upper tail of high-MAE periods\,---\,the conservative-strategy effect of uncertainty-aware planning, avoiding the worst-case publications at the cost of occasionally accepting slightly higher errors on easy periods.
A representative within-period example is shown in \cref{fig:settlement_graph}, illustrating how the uncertainty-aware methods anticipate the final sign of the SI in a period where the average crosses zero late.

\begin{table*}[t]
\centering
\caption{
Linear regime ($\costExponent = 1$): published-price MAE across wall-clock budget settings. Each cell reports the across-period mean $\pm$ the Q1--Q3 inter-quartile interval, computed over 1000 settlement periods.
}
\label{tab:results_linear_q1}
\renewcommand{\arraystretch}{1.15}
\setlength{\tabcolsep}{6pt}
\begin{tabular}{lcccc}
\toprule
\multirow{2}{*}{\textbf{Technique}} &
\multicolumn{4}{c}{\textbf{Wall-clock budget}} \\
\cmidrule(lr){2-5}
& \textbf{1 s} & \textbf{2 s} & \textbf{4 s} & \textbf{8 s} \\
\midrule
\addlinespace[2pt]
Perfect Knowledge
& $15.34 \;\pm\; [7.21,\;23.84]$
& $15.34 \;\pm\; [7.21,\;23.84]$
& $15.34 \;\pm\; [7.21,\;23.84]$
& $15.34 \;\pm\; [7.21,\;23.84]$ \\
\midrule
Rule-Based
& $30.30 \;\pm\; [22.28,\;38.38]$
& $30.30 \;\pm\; [22.28,\;38.38]$
& $30.30 \;\pm\; [22.28,\;38.38]$
& $30.30 \;\pm\; [22.28,\;38.38]$ \\
Stochastic MPC
& $22.81 \;\pm\; [13.52,\;30.14]$
& $22.81 \;\pm\; [13.52,\;30.14]$
& $22.82 \;\pm\; [13.52,\;30.14]$
& $22.81 \;\pm\; [13.52,\;30.14]$ \\
Deterministic MPC
& $32.67 \;\pm\; [20.28,\;44.68]$
& $32.67 \;\pm\; [20.28,\;44.68]$
& $32.67 \;\pm\; [20.28,\;44.68]$
& $32.67 \;\pm\; [20.28,\;44.68]$ \\
\midrule
MCTS
& $27.53 \;\pm\; [22.38,\;32.63]$
& $27.50 \;\pm\; [22.00,\;32.50]$
& $27.49 \;\pm\; [21.89,\;32.33]$
& $27.65 \;\pm\; [21.95,\;32.77]$ \\
\textbf{S3TS (our)}
& $\mathbf{26.32} \;\pm\; [22.30,\;30.93]$
& $\mathbf{26.21} \;\pm\; [22.21,\;30.75]$
& $\mathbf{26.03} \;\pm\; [21.93,\;30.55]$
& $\mathbf{25.93} \;\pm\; [21.92,\;30.37]$ \\
\bottomrule
\end{tabular}
\end{table*}

\begin{table*}
\centering
\setlength{\tabcolsep}{4pt}
\caption{Non-Linear regime ($\costExponent = 1$): published-price MAE across wall-clock budget settings. Each cell reports the across-period mean $\pm$ the Q1--Q3 inter-quartile interval, computed over 1000 settlement periods.}
\renewcommand{\arraystretch}{1.15}
\setlength{\tabcolsep}{6pt}
\begin{tabular}{lcccc}
\toprule
\multirow{2}{*}{\textbf{Technique}} &
\multicolumn{4}{c}{\textbf{Wall-clock budget}} \\
\cmidrule(lr){2-5}
& \textbf{1 s} & \textbf{2 s} & \textbf{4 s} & \textbf{8 s} \\
\midrule
\addlinespace[2pt]
Rule-Based 
& $53.33 \pm [35.00, 70.00]$ 
& $53.33 \pm [35.00, 70.00]$ 
& $53.33 \pm [35.00, 70.00]$ 
& $53.33 \pm [35.00, 70.00]$ \\
MCTS 
& $41.70 \pm [30.67, 50.94]$ 
& $41.71 \pm [30.65, 50.94]$
& $41.78 \pm [30.40, 51.17]$ 
& $41.86 \pm [30.48, 51.25]$ \\
\textbf{S3TS (our)} 
& $\mathbf{40.77} \pm [31.31, 49.00]$ 
& $\mathbf{40.42} \pm [30.98, 48.92]$ 
& $\mathbf{40.48} \pm [30.92, 48.94]$ 
& $\mathbf{40.24} \pm [30.67, 48.58]$ \\
\bottomrule
\end{tabular}
\label{tab:results_non_linear_q1}
\end{table*}

\begin{table*}
\centering
\setlength{\tabcolsep}{4pt}
\caption{Non-Linear regime ($\costExponent = 2$): published-price MSE across wall-clock budget settings. Each cell reports the across-period mean $\pm$ the Q1--Q3 inter-quartile interval, computed over 1000 settlement periods.}
\renewcommand{\arraystretch}{1.15}
\setlength{\tabcolsep}{6pt}
\begin{tabular}{lcccc}
\toprule
\multirow{2}{*}{\textbf{Technique}} &
\multicolumn{4}{c}{\textbf{Wall-clock budget}} \\
\cmidrule(lr){2-5}
& \textbf{1 s} & \textbf{2 s} & \textbf{4 s} & \textbf{8 s} \\
\midrule
\addlinespace[2pt]
Rule-Based & 
$5673.6 \pm [3735.0, 7410.0]$ & 
$5673.6 \pm [3735.0, 7410.0]$ & 
$5673.6 \pm [3735.0, 7410.0]$ & 
$5673.6 \pm [3735.0, 7410.0]$ \\
MCTS & 
$2987.6 \pm [1770.6, 3852.2]$ & 
$2991.3 \pm [1739.5, 3823.0]$ & 
$2932.4 \pm [1687.3, 3775.2]$ & 
$2942.0 \pm [1673.5, 3780.1]$ \\
\textbf{S3TS (our)} & 
$\mathbf{2895.7} \pm [1811.2, 3736.6]$ & 
$\mathbf{2878.3} \pm [1792.0, 3691.0]$ & 
$\mathbf{2826.4} \pm [1790.3, 3604.5]$ & 
$\mathbf{2782.1} \pm [1750.0, 3543.9]$ \\
\bottomrule
\end{tabular}
\label{tab:results_non_linear_q2}
\end{table*}

\subsection{Non-Linear Dynamics with Linear Cost Term}
\label{sec:non-linear_q1_results}

\Cref{tab:results_non_linear_q1} reports the MAE achieved by each technique on the non-linear regime under $\costExponent = 1$. 
The non-linear dynamics make the problem analytically intractable, so the comparison narrows to the three techniques whose machinery does not depend on a closed-form model: Rule-Based, Deterministic-Median MCTS, and S3TS.

Both planning methods remain substantially better than the rule-based baseline\,---\,a $24.5\%$ improvement for S3TS and $21.5\%$ for MCTS at $8$\,s. 
The advantage of advanced planning therefore carries over from the linear regime: the discontinuous price ladder and the non-linear actors' response still allow either tree search method to extract structure that a naive heuristic cannot.
S3TS continues to outperform its deterministic MCTS counterpart at every computational budget, where the difference widens with increasing budget: from $2.2\%$ at $1$\,s to $3.9\%$ at $8$\,s, with S3TS improving monotonically while MCTS stays mostly flat. 
This is the same pattern observed in the linear case, now reproduced under genuinely non-linear dynamics and without an available analytical reference. 
The continued benefit of explicit uncertainty handling under these conditions is the central finding of this experiment: it indicates that the strength of S3TS does not rest on the tractability of the surrounding problem, and that the gains we attributed to scenario-tree reasoning in the linear case can still be reaped for non-smooth dynamics and when the cost function is no longer affine. 

\begin{figure}
    \centering
    \includegraphics[width=0.5\textwidth]{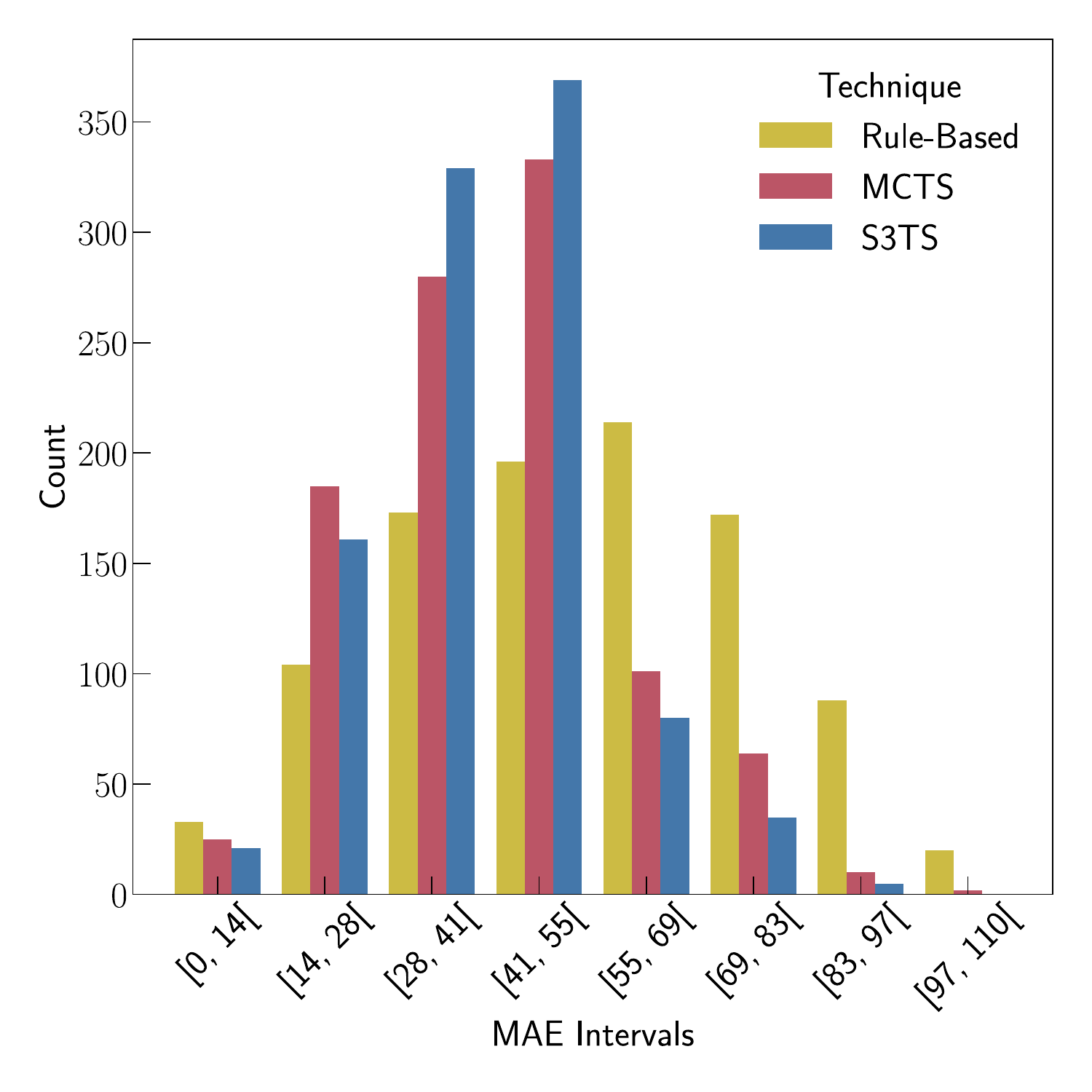}
    \caption{Publication MAE histogram in the non-linear regime with $\costExponent \doteq 1$.}
    \label{fig:non_linear_q1_histogram}
\end{figure}

The distributional view in \cref{fig:non_linear_q1_histogram} adds qualitative support to our observation for \cref{tab:results_non_linear_q1}. The Rule-Based's MAE distribution is broadly spread
across the MAE axis, with substantial mass in the high-error bins $[55, 83[$ and a non-negligible tail extending beyond MAE $80$. 
This spread reflects the heuristic's vulnerability to the discontinuous price ladder: whenever late-period SI dynamics push the final price across a discontinuity that the running average failed to anticipate, the heuristic incurs a large publication error. Both tree search methods relocate the distribution toward lower errors, but they do so in different ways. 
S3TS concentrates its mass tightly in the central $[28, 55[$ range, accepting consistent mid-range errors in exchange for sharply suppressing the high-error tails. MCTS, by contrast, spreads more broadly: while it places more periods in the low-error bin $[14, 28[$ than S3TS does, it also retains noticeably more mass in the $[55, 83[$ tails. 
The conservative strategy of S3TS therefore manifests not only as an average reduction in error, but also as a deliberate trade\,---\,getting fewer exceptionally good publications, but also fewer exceptionally bad ones.

\subsection{Non-Linear Dynamics with Quadratic Cost Term}
\label{sec:non-linear_q2_results}

\Cref{tab:results_non_linear_q2} reports the MSE achieved by each technique on the non-linear regime under the quadratic cost exponent ($\costExponent=2$). 
The qualitative story carries over from the $\costExponent = 1$ case: both tree search methods substantially outperform Rule-Based, S3TS outperforms Deterministic-Median MCTS at every budget, and the S3TS--MCTS gap widens with computational budget (from $3.1\%$ at $1$\,s to $5.4\%$ at $8$\,s, with S3TS improving monotonically while MCTS remains mostly flat).

\begin{figure}
    \centering
    \includegraphics[width=0.5\textwidth]{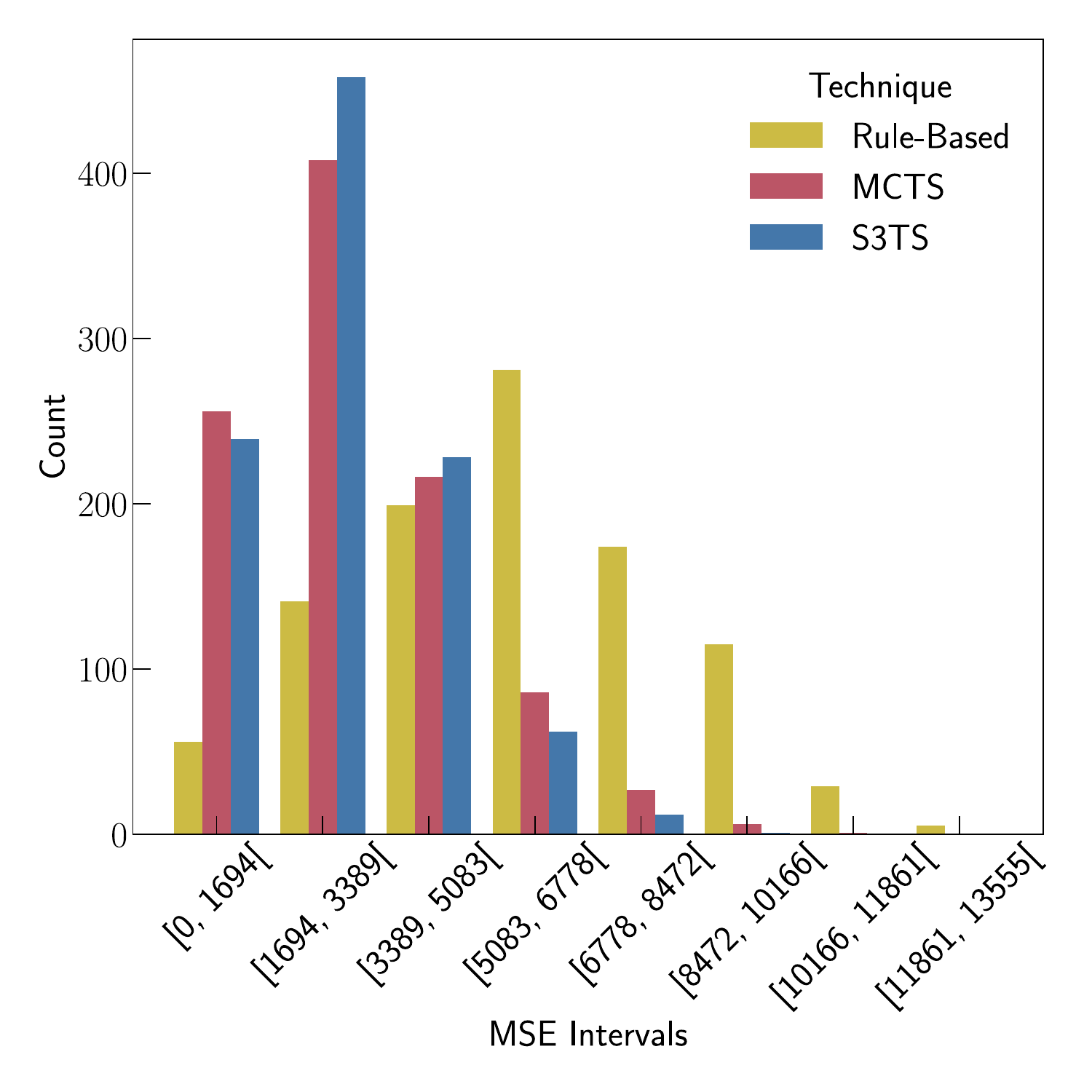}
    \caption{Publication MSE histogram in the non-linear regime with $\costExponent \doteq 2$.}
    \label{fig:non_linear_q2_histogram}
\end{figure}

What changes under the quadratic cost is the magnitude of the advantages. 
The gap between Rule-Based and the tree search methods roughly doubles, from $24.5\%$ under MAE to $51.0\%$ under MSE.
This is expected from the form of the cost function: by squaring the publication error, MSE places disproportionate weight on the high-error tail of the per-period distribution, and as the histogram in \cref{fig:non_linear_q2_histogram} shows, the Rule-Based distribution extends substantially further into that tail than either tree search method. 
The S3TS--MCTS gap also widens slightly under MSE\,---\,from $3.9\%$ in the $\costExponent=1$ experiment to $5.4\%$ with $\costExponent=2$ at the largest budget\,---\,for the same reason: S3TS's conservative-strategy effect suppresses moderate-to-high errors more aggressively than MCTS does, and that suppression matters more when errors are weighted quadratically.

The distributional view in \cref{fig:non_linear_q2_histogram} makes this concrete.
The histogram exhibits the same risk-trading pattern already discussed in \cref{sec:non-linear_q1_results}, now amplified by the quadratic cost: S3TS concentrates its mass tightly around the modal bin, MCTS retains visibly more mass in both the low-error and the high-error bins, and Rule-Based spreads across the full range with a thin tail at the high end. 
The difference is one of emphasis rather than kind: under quadratic cost, the bins on the high-error side of the modal bin contribute disproportionately to the mean, so the portion of the MCTS distribution sitting in those bins weighs more heavily here than the equivalent portion did under MAE. The $5.4\%$ S3TS--MCTS gap at $8$\,s reflects almost entirely this tail asymmetry.

\section{Conclusions and Future Directions}
\label{sec:conclusions}

We introduced S3TS, an adaptation of MCTS to multi-stage stochastic optimization through scenario trees. S3TS inherits the flexibility of the MCTS framework with respect to model complexity, supporting non-linear dynamics and non-linear cost functions, which both fall outside the capabilities of conventional mathematical-programming solvers. 
At the same time, by taking a pre-built scenario tree as input rather than sampling outcomes online or learning them through a latent model (as in most popular stochastic MCTS alternatives), it aligns with the multi-stage stochastic optimization tradition and can be driven by the same scenario-generation pipelines that already feed stochastic MPC.

We evaluated S3TS on a demand response signal publication problem, inspired by the imbalance settlement mechanism currently in place in Belgium. The evaluation spanned three experiments and a matched wall-clock budget protocol of $1$, $2$, $4$, and $8$ seconds per publication.\footnote{The computational budgets considered allow for realistic applications, where publications typically happen at a minute pace.} In the linear regime (\ie with simplified linear dynamics), S3TS came within $14\%$ of the optimum of the multi-stage stochastic problem solved with MILP\,---\,a proof of concept that the algorithm closely tracks the stochastic optimum even without direct access to the analytical model. 
In the two non-linear experiments (with linear and quadratic cost terms), where no MILP reference is available, S3TS outperformed both a rule-based baseline and Deterministic-Median MCTS at every budget. 
Two findings recurred across all three experiments and deserve emphasis: the S3TS--MCTS gap \emph{widens} rather than narrows with budget, indicating that the benefit of explicit uncertainty representation grows with available compute rather than diminishing; and the distributional pattern is consistent across regimes, with S3TS trading a small number of exceptionally low-error publications for substantially fewer high-error ones\,---\,a conservative-strategy effect that pays off particularly under quadratic cost.

Several avenues remain to enhance the algorithm. 
The most immediate is to enrich the search phase with prior knowledge, in the spirit of \cite{silver2017mastering}: value estimates could be attached to state nodes, but also\,---\,perhaps more interestingly\,---\,to \textualScenarioNode nodes, allowing the tree search to directly compute toward the genuinely informative branches. 
This could also help close the matched-budget gap to Stochastic MPC observed in the linear regime, since much of the residual $14\%$ is plausibly attributable to the simulation budget rather than to a structural limitation of the algorithm. A second direction is to evaluate S3TS across a more diverse set of problem domains; the demand response problem considered here was chosen as a tractable proof-of-concept, but the algorithm's flexibility suggests it could be equally applied to residential heating, microgrid scheduling, and complex maintenance optimization, among others. 
Finally, the algorithm's objective function allows for a straightforward extension to risk-sensitive criteria such as CVaR, which would let it explicitly
target the tail behavior that the present quadratic-cost results already show to be the dominant differentiator of S3TS compared to conventional MCTS.

\section*{Acknowledgments}
This research was partly funded by the Flemish Government through
the “Onderzoeksprogramma Artificiële Intelligentie (AI) Vlaanderen” programme as well as a travel grant from the Research Foundation - Flanders, by the Energy Transition Fund

\printcredits

\bibliographystyle{cas-model2-names}

\bibliography{cas-refs}

\end{document}